\documentclass{article}

\usepackage{microtype}
\usepackage{graphicx}
\usepackage{subcaption}
\usepackage{booktabs} 

\usepackage{hyperref}

\usepackage[accepted]{icml2026}
\makeatletter
\renewcommand{\Notice@String}{}
\makeatother

\usepackage{amsmath}
\usepackage{amssymb}
\usepackage{mathtools}
\usepackage{amsthm}
\usepackage{multirow}
\usepackage{pifont}
\newcommand{\cmark}{\ding{51}} 
\newcommand{\xmark}{\ding{55}} 
\usepackage{booktabs}
\usepackage[table]{xcolor}   
\usepackage{array}
\usepackage{bm}
\usepackage{algorithm}

\usepackage{algpseudocode} 
\usepackage{amsmath}
\makeatletter

\usepackage{makecell}

\AtBeginDocument{%
  \setlength{\abovedisplayskip}{4pt}%
  \setlength{\belowdisplayskip}{4pt}%
  \setlength{\abovedisplayshortskip}{4pt}%
  \setlength{\belowdisplayshortskip}{4pt}%
}

\newcommand{\best}[1]{\textbf{#1}}

\usepackage[capitalize,noabbrev]{cleveref}

\newlength{\myparskip}
\theoremstyle{plain}

\theoremstyle{definition}

\theoremstyle{remark}

\usepackage[textsize=tiny]{todonotes}

\icmltitlerunning{MedScope: Incentivizing ``Think with Videos" for Clinical Reasoning via Coarse-to-Fine Tool Calling}

\begin{document}

\twocolumn[
  \icmltitle{MedScope: Incentivizing ``Think with Videos" for Clinical Reasoning \\ via Coarse-to-Fine Tool Calling}



  \icmlsetsymbol{equal}{*}

  \begin{icmlauthorlist}
    \icmlauthor{Wenjie Li}{equal,a,c}
    \icmlauthor{Yujie Zhang}{equal,c,b}
    \icmlauthor{Haoran Sun}{b,e}
    \icmlauthor{Xingqi He}{b}
    \icmlauthor{Hongcheng Gao}{d}
    \icmlauthor{Chenglong Ma}{c,b}
    \icmlauthor{Ming Hu}{e}
    \icmlauthor{Guankun Wang}{f}
    \icmlauthor{Shiyi Yao}{g}
    \icmlauthor{Renhao Yang}{g}
    \icmlauthor{Hongliang Ren}{f}
    \icmlauthor{Lei Wang}{g}
    \icmlauthor{Junjun He}{c,e}
    \icmlauthor{Yankai Jiang}{e}
  \end{icmlauthorlist}

  \icmlaffiliation{a}{College of Health Science and Technology, Shanghai Jiao Tong University School of Medicine, Shanghai, China}
  \icmlaffiliation{b}{Fudan University, Shanghai, China}
  \icmlaffiliation{c}{Shanghai Innovation Institute, Shanghai, China}
  \icmlaffiliation{d}{Tsinghua University, Beijing, China}
  \icmlaffiliation{e}{Shanghai Artificial Intelligence Laboratory, Shanghai, China}
  \icmlaffiliation{f}{The Chinese University of Hong Kong, Hong Kong, China}
  \icmlaffiliation{g}{Ruijin Hospital, Shanghai Jiaotong University, Shanghai, China}

  \icmlcorrespondingauthor{Yaikai Jiang}{jiangyankai@pjlab.org.cn}
  \icmlcorrespondingauthor{Junjun He}{hejunjun@sjtu.edu.cn}
  \icmlcorrespondingauthor{Lei Wang}{ray\_wangs@hotmail.com}


  \vskip 0.3in
]



\printAffiliationsAndNotice{\icmlEqualContribution}
\enlargethispage{2\baselineskip} 

    \begin{abstract}
    Long-form clinical videos are central to visual evidence-based decision-making, with growing importance for applications such as surgical robotics and related settings. However, current multimodal large language models typically process videos with passive sampling or weakly grounded inspection, which limits their ability to iteratively locate, verify, and justify predictions with temporally targeted evidence. To close this gap, we propose \textbf{MedScope}, a tool-using clinical video reasoning model that performs coarse-to-fine evidence seeking over long-form procedures. By interleaving intermediate reasoning with targeted tool calls and verification on retrieved observations, MedScope produces more accurate and trustworthy predictions that are explicitly grounded in temporally localized visual evidence. To address the lack of high-fidelity supervision, we build \textbf{ClinVideoSuite}, an evidence-centric, fine-grained clinical video suite. We then optimize \textbf{MedScope} with \textbf{G}rounding-\textbf{A}ware \textbf{G}roup \textbf{R}elative \textbf{P}olicy \textbf{O}ptimization (\textbf{GA-GRPO}), which directly reinforces tool use with grounding-aligned rewards and evidence-weighted advantages. On full and fine-grained video understanding benchmarks, \textbf{MedScope} achieves state-of-the-art performance in both in-domain and out-of-domain evaluations. Our approach illuminates a path toward medical AI agents that can genuinely “think with videos” through tool-integrated reasoning. We will release our code, models, and data.
    \end{abstract}
\vspace{-2.0em}

\setlength{\parskip}{\myparskip}

\section{Introduction}

\begin{figure}[t]
    \centering
    \includegraphics[width=\columnwidth]{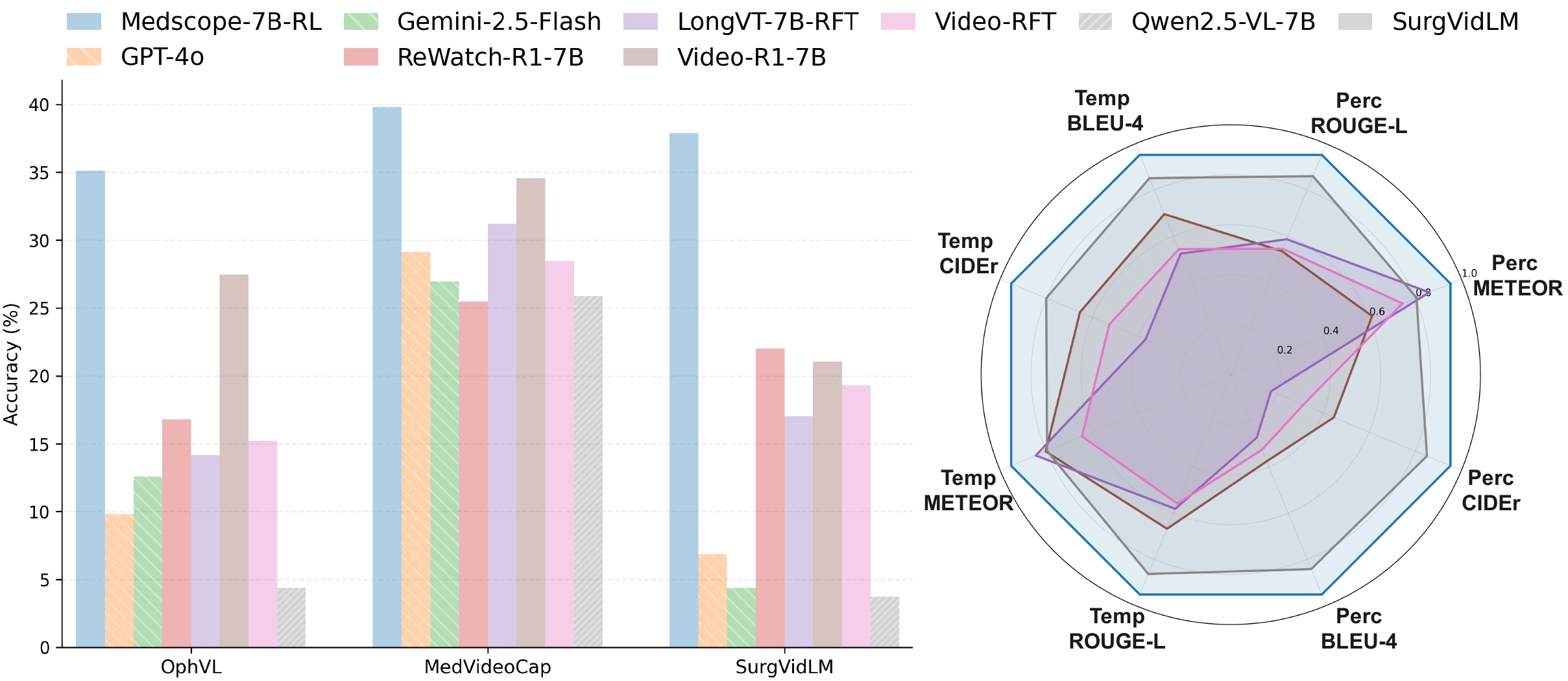}
    \caption{\textbf{Performance comparison on full and fine-grained video understanding and VQA benchmarks.} Left: grounded VQA accuracy on ClinVideo-Eval. Right: full and fine-grained video understanding quality on SVU-31K.}
    \vspace{-12pt}
    \label{fig:intro_result}
\end{figure}

\begin{figure*}[t]
    \centering
    \includegraphics[width=0.92\linewidth]{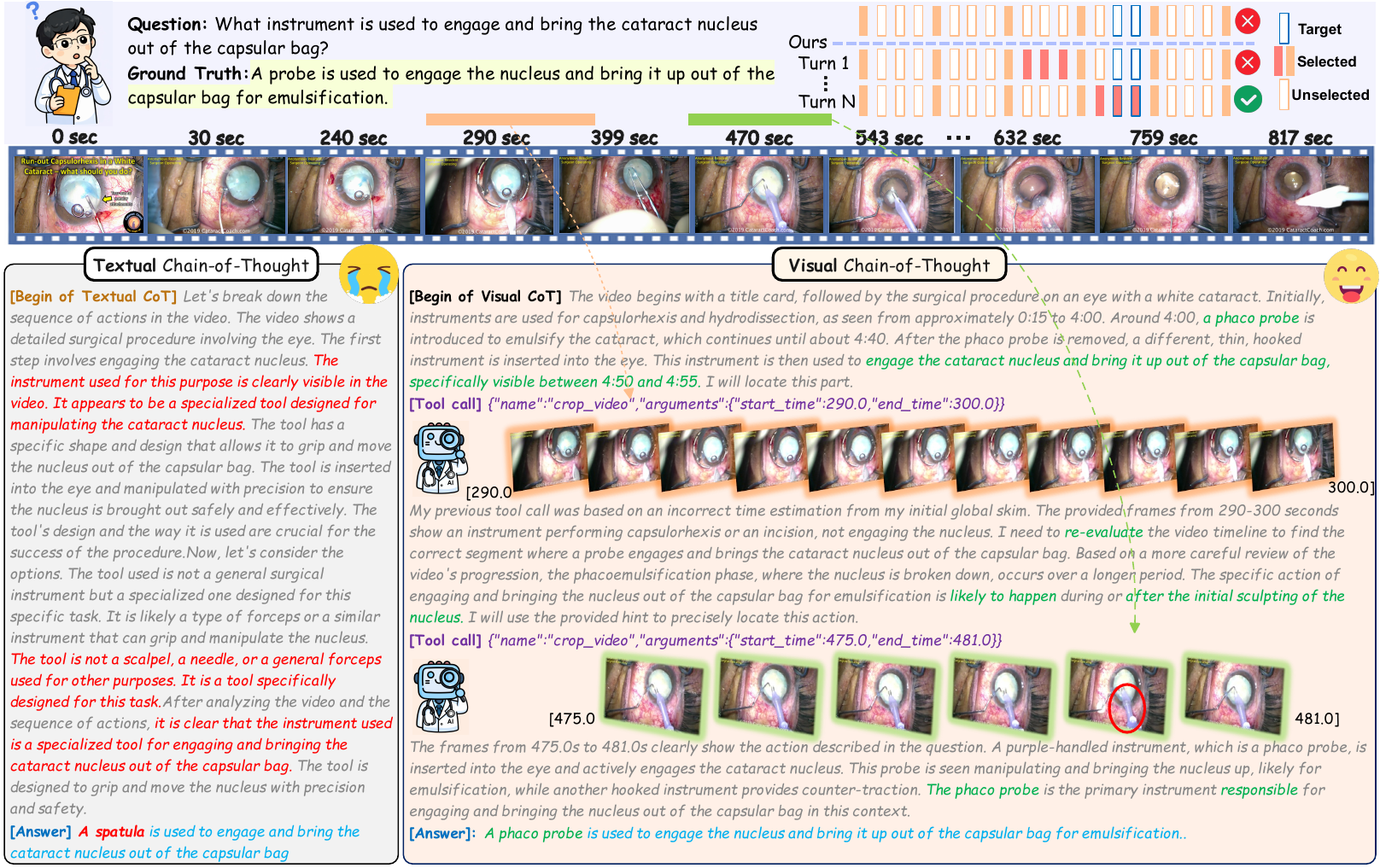}
    \vspace{-3pt}
    \caption{\textbf{Comparison between textual CoT and visual CoT for evidence-grounded clinical video reasoning.} Left: textual CoT shows overconfident hallucinations (red), inventing rationales and predicting the wrong instrument. Right: visual CoT iteratively retrieves and integrates dense visual evidence via tool calls, grounding reasoning in localized observations and producing the correct answer.}
    \vspace{-12pt}
    \label{fig:intro}
\end{figure*}
Understanding long videos that span tens of minutes remains challenging for multimodal intelligence, because decisive yet temporally sparse evidence is buried within complex event sequences across thousands of frames~\citep{qian2024streaming,liu2025longvideoagent,ma2025drvideo,he2024ma,tang2025video}. In medicine, long surgical and endoscopic recordings support intraoperative decision making\citep{biffi2022novel}, postoperative review\citep{al2024impact}, and surgical training\citep{gastager2025watch}, where conclusions often hinge on brief, fine-grained cues and must be grounded in concrete visual evidence\citep{kiyasseh2023vision}. Clinicians therefore do not watch an entire procedure in one pass; they first skim the global course to form hypotheses, then revisit short temporal windows to retrieve and verify subtle signs\citep{van2024your}.

Recent advances in multimodal chain-of-thought (CoT) reasoning have improved interpretability, and large multimodal models (LMMs) perform strongly on short-video understanding\citep{zhang2025improve,wang2025multimodal}. Yet most methods still follow an R1-style\citep{guo2025deepseek}, text-first pipeline that treats the visual stream as static context and does not support explicit evidence seeking or iterative verification\citep{feng2025video,li2025videochat,liu2025visual}. This design is fragile for long videos, where sparse sampling can miss decisive moments and encourage plausible but unsupported rationales\citep{yao2025k}. This raises a key question: can LMMs learn to search and verify evidence over time for long-form medical videos?

We answer this question with an interleaved visual CoT (VCoT) framework that moves beyond text-centric reasoning by enabling LMMs to couple hypothesis-driven reasoning with on-demand, coarse-to-fine temporal localization and iterative verification in long clinical videos\citep{yang2025longvt,zhang2025rewatch}. Inspired by how clinicians skim globally and then revisit locally, the model first forms a coarse understanding and then dynamically selects where to look next, repeatedly re-inspecting candidate moments and revising its hypothesis when needed. As shown in Figure~\ref{fig:intro}, this behavior arises from the LMM’s native temporal grounding ability and is realized through a lightweight visual toolbox that supports both clip-level densification and frame-level verification\citep{ge2025framemind,he2025framethinker}, enabling iterative evidence seeking rather than one-pass inference. In practice, this paradigm strengthens LMMs' long-video understanding, as evidenced in Figure~\ref{fig:intro_result}.

In this paper, we present \textbf{MedScope}, a tool-using clinical video reasoning model that enables ``think with videos'' through iterative evidence seeking and verification over long-form procedures. However, existing medical video datasets still lack dense evidence-linked supervision and standardized evaluation for verification-driven reasoning. To bridge this gap, we introduce \textbf{ClinVideoSuite}, a large-scale evidence-centric suite with grounded QA and tool-augmented trajectories, including \textbf{ClinVideo-QA-254K} and \textbf{ClinVideo-VCoT-34K}, plus caption and CoT subsets. Building on ClinVideoSuite, we train MedScope in three stages to elicit doctor-like coarse-to-fine verification. We first perform clinical reasoning warm-up to learn medical semantics and long-horizon reasoning. We then conduct visual CoT cold-start supervised fine-tuning (SFT) to teach when additional evidence is needed and how to acquire it via native tools for clip-level densification and frame-level verification. Finally, we propose \textbf{GA-GRPO}, an agentic RL recipe with a grounding-aware reward and adaptive advantage reweighting to encourage temporally aligned tool use and stable learning under diverse video conditions.

Our contributions are summarized as follows:

\begin{itemize}
    \item We propose a medical visual CoT paradigm for ``thinking with videos", and instantiate it with \textbf{MedScope}, a tool-using video LMM trained via a three-stage pipeline that elicits coarse-to-fine clinical reasoning.
   \item We build an evidence-centric data suite, \textbf{ClinVideoSuite}, for training and evaluation, including \textbf{ClinVideo-eval}, a fine-grained long-form medical video reasoning benchmark with traceable evidence.
    \item We introduce \textbf{GA-GRPO}, an agentic RL recipe with grounding-aware rewards and adaptive advantages to promote temporally aligned tool use.
    \item MedScope achieves \textbf{state-of-the-art} performance among \textbf{open-source} models on standard medical video benchmarks under both in-domain and out-of-domain evaluations.
\end{itemize}

\section{Related Work}
\subsection{Long-Video Reasoning in MLLMs}
\label{Long-Video Reasoning in MLLMs}
Reasoning over long videos remains challenging for multimodal large language models (MLLMs) due to high computation and sparse task-relevant evidence~\citep{zou2024seconds, tang2025video}. Iterative temporal selection methods such as MIST~\citep{gao2023mist} and SeViLA~\citep{yu2023self} improve efficiency by localizing segments before answering, but largely decouple localization from reasoning. Long-context MLLMs including LLaMA-VID~\citep{li2024llama}, Flash-VStream~\citep{zhang2024flash}, LongVA~\citep{zhang2024long}, and LongVILA~\citep{chen2024longvila} scale via token compression or long-context modeling, yet still rely on passive sampling and remain costly. Coarse-to-fine sampling strategies~\citep{jeoung2024adaptive,yao2025generative} further improve efficiency by selecting informative frames on demand. Additionally, reinforcement learning has been explored for video reasoning (e.g., Video-R1~\citep{feng2025video}, VideoChat-R1~\citep{li2025videochat}, LongVILA-R1~\citep{chen2025scaling}), but these methods often use outcome-level rewards under weak supervision, which can misalign intermediate reasoning with visual evidence. 
In contrast, our work integrates agentic interaction with RL to tightly couple coarse-to-fine visual exploration with reasoning for efficient, grounded medical long-video understanding.

\subsection{Tool-Augmented Medical MLLMs}
\label{Tool-Augmented Medical MLLMs}
Augmenting medical MLLMs with external tools is an effective way to extend standalone reasoning by leveraging specialized perception modules and expert systems. Early works such as MMedAgent~\citep{li2024mmedagent} and VILA-M3~\citep{nath2025vila} train models to invoke tools for tasks like segmentation and classification, while AURA~\citep{fathi2025aura} unifies multiple domain-specific models, including MedSAM~\citep{ma2024segment} and CheXAgent~\citep{chen2024chexagent}, in a single tool-augmented framework. Despite strong task-level performance, these systems often rely on fragmented tool calls that limit coherent planning and reflective reasoning. To better support multi-step clinical decision-making, MedAgent-Pro~\citep{wang2025medagent} and SMR-Agents~\citep{wang2026smr} adopt hierarchical or collaborative multi-agent designs, but their dependence on predefined workflows and prompt-level orchestration can hinder generalization and robustness to tool failures. More recently, inspired by the ``thinking with images'' paradigm, iterative tool-mediated visual reasoning has been shown to improve grounding and reduce hallucinations~\citep{jiang2025incentivizing}. However, existing tool-augmented medical MLLMs mainly focus on static images and do not address the dynamic visual exploration and evidence accumulation required for long-horizon medical video reasoning~\citep{jiang2026ibisagent}.
In this work, we move beyond image-centric tool use and learn tool-mediated, temporally adaptive evidence gathering for clinical long-video reasoning.

\section{Method}

\begin{figure*}[t]
    \centering
    \includegraphics[width=0.92\linewidth]{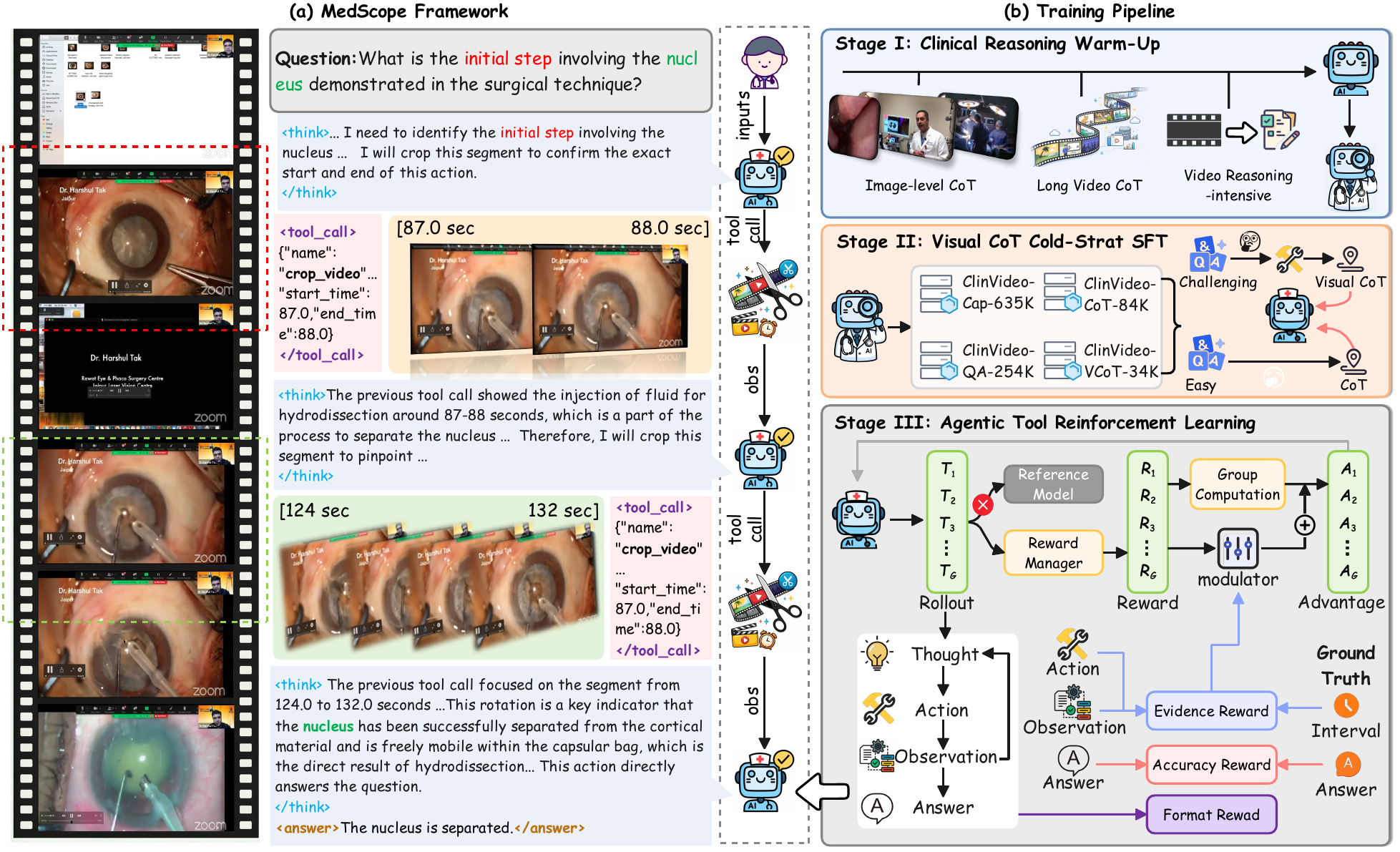}
    \vspace{-3pt}
    \caption{\textbf{Overview of MedScope.} (a) Coarse-to-fine clinical reasoning with explicit thought and tool actions that progressively retrieve temporally targeted dense evidence for verification. (b) Three-stage training pipeline: clinical reasoning warm-up, visual-CoT cold-start SFT on ClinVideoSuite, and agentic tool reinforcement learning with grounding-aware rewards and advantage shaping.
}
    \vspace{-12pt}
    \label{fig:method}
\end{figure*}

\subsection{Overview}

MedScope enables ``thinking with videos'' via coarse-to-fine clinical reasoning with a lightweight visual toolbox, alternating textual reasoning and tool-based evidence acquisition to verify predictions with temporally targeted dense observations (Section~\ref{sec:coarse_to_fine_reasoning_framework}). To support this behavior, we build ClinVideoSuite with evidence-centric captions, QA pairs tied to localized supervision windows, and environment-interactive visual-CoT trajectories produced by native tools (Section~\ref{sec:clinvideosuite}). We then train the model in three stages: warm-up, visual-CoT cold-start, and grounding-aware GRPO, so that tool use is both effective and evidence-grounded (Section~\ref{sec:training_strategy}).

\subsection{Coarse-to-Fine Clinical Reasoning Framework}
\label{sec:coarse_to_fine_reasoning_framework}
As shown in Figure~\ref{fig:method}(a), MedScope performs coarse-to-fine clinical video reasoning by alternating between hypothesis-driven reasoning and tool-based evidence retrieval, producing an explicit trajectory for verification.

\paragraph{Textual and visual rationalization.}
We decouple reasoning from evidence acquisition. At round $k$, the model outputs a textual rationale $t_{i,k}$ that states its current belief and the evidence it seeks, then invokes an action $a_{i,k}$ to retrieve an observation $o_{i,k}$ as visual evidence. This structured separation makes the decision process inspectable and supports iterative coarse-to-fine verification.
\vspace{-5pt}

\paragraph{Multi-round generation and trajectory.}
Given a query $q_i$ and video $v_i$, the model generates a multi-round trajectory:
\begin{equation}
\tau_i=\{(t_{i,k},a_{i,k},o_{i,k})\}_{k=1}^{K_i},
\end{equation}
where $t_{i,k}$ denotes the textual rationale, $a_{i,k}$ the selected tool action, and $o_{i,k}$ the returned visual evidence. At each round, the policy predicts $(t_{i,k},a_{i,k})=\pi_\theta(c_{i,k})$ conditioned on the accumulated context $c_{i,k}$. If $a_{i,k}$ invokes a tool, the environment returns $o_{i,k}=\mathcal{E}(v_i,a_{i,k})$, and we update the context as $c_{i,k+1}=c_{i,k}\oplus(t_{i,k},a_{i,k},o_{i,k})$ for the next step. We enforce a structured interface within each trajectory: \texttt{<think>} for textual rationalization, \texttt{<tool\_call>} for visual rationalization, \texttt{<tool\_response>} for tool observations, and \texttt{<answer>} for the terminal prediction.

\paragraph{Visual Toolbox}
More complex and diverse examples of the model’s reasoning process are provided in Appendix~\ref{appendix:case}. In this paper, the action space includes two native tools that support hierarchical inspection:
\begin{itemize}
  \item \texttt{crop\_video}: takes a video and a time window, and returns a clip-level densely sampled frame sequence within the window for localized evidence gathering.
  \item \texttt{get\_frame}: takes a video and a timestamp, and returns the three frames nearest to that timestamp for fine-grained verification.
\end{itemize}
\vspace{-3pt}

\subsection{ClinVideoSuite: Evidence-Centric Data Synthesis}
\label{sec:clinvideosuite}

\begin{figure*}[t]
    \centering
    \includegraphics[width=1\linewidth]{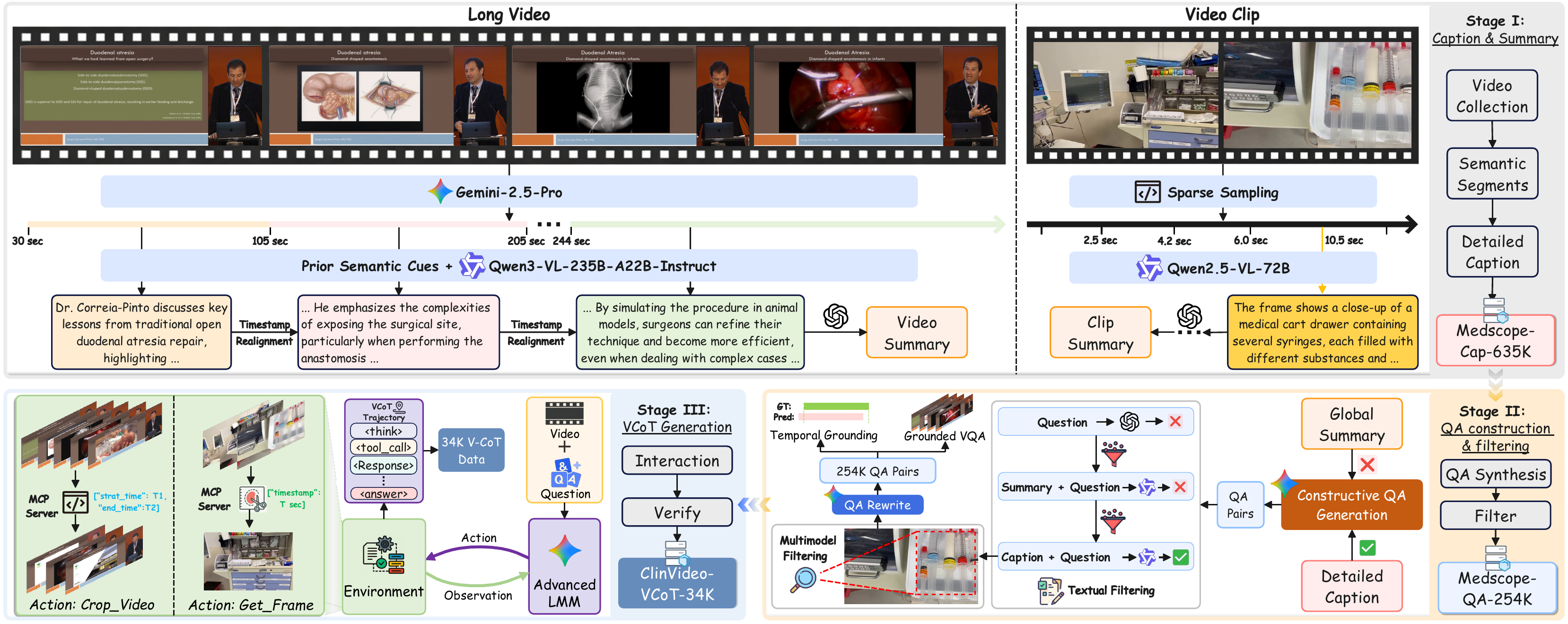}
    \caption{\textbf{ClinVideoSuite data synthesis pipeline.} Stage 1 builds evidence-centric dense captions and global summaries. Stage 2 generates and filters QA with text checks and multimodal verification to enforce localized evidence dependence. Stage 3 collects tool-augmented visual CoT trajectories via native tool interaction in a real video environment.}
    \vspace{-15pt}
    \label{fig:clinvideosuite_pipeline}
\end{figure*}

To address the data bottleneck for think-with-videos in clinical videos, we introduce \textbf{ClinVideoSuite}, a large-scale, high-fidelity, and video-grounded suite that emphasizes dense, temporally localized visual cues for videoQA and temporal grounding (Figure~\ref{fig:clinvideosuite_pipeline}). Built from MedVideoCap~\citep{wang2025medgen}, OphVL~\citep{hu2025ophclip}, and SurgVidLM~\citep{wang2025surgvidlm} via captioning, high-difficulty QA construction, and rationalization synthesis, ClinVideoSuite comprises \textbf{ClinVideo-Cap}, \textbf{ClinVideo-QA}, \textbf{ClinVideo-VCoT}, \textbf{ClinVideo-CoT}, and \textbf{ClinVideo-RL}, which together support training and evaluation of grounded clinical video reasoning.

\paragraph{\textbf{Stage 1: Evidence-centric captioning and global summarization.}}
Given a video $V$ with duration $T$, we construct timestamped dense captions $\mathcal{C}(V)$ as localized evidence and a coarse global summary $S(V)$ as context. We first sparsify $V$ by low-rate sampling $\tilde{V}=\mathcal{D}_{\rho}(V)$, where $\mathcal{D}_{\rho}$ samples at rate $\rho$. A segmenter $\Phi$ then predicts an entity-guided partition:
\begin{equation}
\mathcal{B}(V)=\Phi(\tilde{V})=\{(e_m,s_m,u_m)\}_{m=1}^{M},
\end{equation}
where $e_m$ is an entity cue and $[s_m,u_m]\subseteq[0,T]$ is the associated window. We densify each window to generate fine-grained descriptions, forming $\mathcal{C}(V)=\{(d,t)\}$ with $t\in[0,T]$, and summarize temporally ordered captions into $S(V)=\mathrm{Sum}(\mathrm{Merge}(\{d\}))$, which preserves global context while leaving localized cues to be retrieved from $\mathcal{C}(V)$ for downstream grounding.

\paragraph{\textbf{Stage 2: High-quality QA construction with cross-model filtering.}}
In this stage, we construct open-ended QA pairs that hinge on temporally localized visual cues. For each window $w=[s,e]$, we derive a local view $\mathcal{C}(V;w)$ from dense captions and prompt a generator to produce candidate pairs $(q,a)$ that target details in $\mathcal{C}(V;w)$ but absent from the global summary $S(V)$, while inheriting the same supervision window $w$.

\paragraph{Three-layer text filtering.}
Raw candidates pass a voting-based filtering cascade designed to remove knowledge-driven questions, summary-solvable shortcuts, and internally inconsistent pairs. Let $\mathcal{M}$ be a pool of LLMs, and let $\mathrm{Judge}(\cdot,\cdot)$ judge whether two answers are equivalent. We define a consensus score under a context view $X$:
\begin{equation}
p_X(q,a)=\frac{1}{|\mathcal{M}|}\sum_{M\in\mathcal{M}}\mathbf{1}\!\left[\mathrm{Judge}\!\left(\mathrm{Ans}(M,q,X),a\right)\right],
\end{equation}
where $\mathrm{Ans}(M,q,X)$ is model $M$ answering $q$ given $X$. Filter A removes knowledge-intensive or guessable items by requiring low text-only consensus, $p_{\varnothing}(q,a) < \theta_{\text{text}}$.
Filter B removes global-summary shortcuts by requiring low consensus given only $S(V)$, $p_{S(V)}(q,a) < \theta_{\text{sum}}$.
Filter C enforces local-evidence dependence by requiring high consensus under local dense captions $\mathcal{C}(V;w)$, $p_{\mathcal{C}(V;w)}(q,a) \ge \theta_{\text{loc}}$.

\paragraph{Multimodal confirmation.}
To eliminate caption hallucinations and ensure genuine video grounding, we perform a final multimodal check that answers $q$ directly from the video segment $\mathrm{Clip}(V,w)$ and retains only pairs with strong agreement against $a$. The surviving items form ClinVideo-QA-254K, and their associated dense evidence supports downstream visual rationalization synthesis.

\paragraph{\textbf{Stage 3: Environment-interactive visual-CoT synthesis.}}
We synthesize visual-CoT trajectories via real interaction with the video environment. A teacher model starts from sparse global context and decides whether to query native tools for temporally targeted observations, then continues reasoning with the observation; otherwise it answers directly. This yields both tool-augmented and tool-free trajectories, improving synthesis efficiency while retaining evidence-seeking behaviors. Prompting details are in Appendix~\ref{appendix:prompts}.

\paragraph{\textbf{Dataset Statistics.}}
We summarize statistics of \textbf{ClinVideo-Cap-635K} and \textbf{ClinVideo-QA-254K}, which form the core of ClinVideoSuite, while remaining subsets are derived from \textbf{ClinVideo-QA-254K} under different reasoning and interaction settings. As shown in Appendix Figures~\ref{fig:data1}--\ref{fig:data2}, \textbf{ClinVideo-Cap-635K} provides large-scale timestamped dense captions with diverse temporal densities across sources, and \textbf{ClinVideo-QA-254K} offers concise, evidence-grounded QA pairs with localized supervision windows. We provide detailed analyses and visualizations in Appendix~\ref{appendix:Dataset Statistics}.

\subsection{Training Strategy}
\label{sec:training_strategy}

As shown in Figure~\ref{fig:method}(b), we adopt a three-stage pipeline for grounded clinical video reasoning: (i) warm up to learn medical semantics and long-horizon reasoning, (ii) tool-augmented visual-CoT cold start to learn when and how to invoke tools, and (iii) GA-GRPO agentic RL to refine tool use via grounding-aware rewards and advantage modulation.

\paragraph{Clinical Reasoning Warm-Up}
We warm up the backbone with multi-source SFT to build general multimodal reasoning, long-video understanding, and surgical visual semantics before introducing tool use. Specifically, we merge (i) Video-R1 video CoT for step-by-step multimodal inference~\citep{feng2025video}, (ii) a curated subset of LongVideo-Reason for long-horizon temporal reasoning~\citep{chen2025scaling}, and (iii) surgical image-level VQA from EndoVis-18-VQA~\citep{bai2024endovis18vqla}, PitVQA~\citep{he2024pitvqa}, and Surgical-VQA~\citep{seenivasan2022surgical}. For the image-level data, we apply rejection sampling to retain only format-compliant, visually grounded rationales. Detailed training objectives are provided in Appendix~\ref{appendix:warmup_sft_objective}.

\paragraph{Visual CoT Cold-Start Supervised Fine-Tuning}
We then perform a cold-start SFT on our visual-CoT data \textbf{ClinVideo-VCoT-34K} and \textbf{ClinVideo-CoT-84K} to teach coarse-to-fine clinical verification in videos. This stage mixes tool-augmented trajectories that retrieve evidence from clips via \texttt{get\_frame} and from long videos via \texttt{crop\_video} with tool-free trajectories for cases that are already resolvable from sparse frames. Training on this mixture promotes adaptive reasoning, helping the model learn when additional evidence is necessary and how to incorporate observations into a faithful rationale.

\paragraph{Grounding-Aware GRPO Training}
We further apply agentic RL to extend coarse-to-fine clinical verification beyond supervised traces. 
Specifically, we instantiate GRPO with tool-aligned grounding rewards, reinforcing rollouts that are correct and temporally well-localized.

\paragraph{Reward Design.}
We use a composite reward $R = R_{\text{acc}} + R_{\text{format}} + R_{\text{evidence}}$, where $R_{\text{acc}}$ scores answer correctness via an LLM judge and $R_{\text{format}}$ checks the required structured interface. The evidence term $R_{\text{evidence}}$ evaluates grounding based on the final tool call, using temporal overlap for \texttt{crop\_video} and timestamp alignment for \texttt{get\_frame}. For \texttt{crop\_video}, we additionally apply an IoU-based bonus schedule on top of the base overlap reward, which further encourages precise temporal alignment as IoU increases. Algorithm~\ref{alg:iou_reward} summarizes the procedure, and the full mathematical definition is deferred to Appendix~\ref{app:reward_details}.

\vspace{-6pt}
\begin{algorithm}[htbp]
\caption{Grounding-Aware Tool Reward}
\label{alg:iou_reward}
\begin{algorithmic}
\Require Prediction $y$, GT $[g_s,g_e]$ or $g_f$, tool calls $T$
\Ensure $R = R_{\text{acc}} + R_{\text{format}} + R_{\text{evidence}}$
\State $R_{\text{acc}} \leftarrow \text{Judge}(y)$
\State $R_{\text{format}} \leftarrow \text{FormatOK}(y)$
\State $R_{\text{evidence}} \leftarrow 0$
\If{$\text{crop\_video}\in T$}
  \State $(p_s,p_e) \leftarrow \text{LastCrop}(T)$
  \State $r_c \leftarrow \text{CropReward}(p_s,p_e)$
  \State $R_{\text{evidence}} \leftarrow R_{\text{evidence}} + r_c$
\EndIf
\If{$\text{get\_frame}\in T$}
  \State $t \leftarrow \text{LastFrame}(T)$
  \State $r_f \leftarrow \text{FrameReward}(t)$
  \State $R_{\text{evidence}} \leftarrow R_{\text{evidence}} + r_f$
\EndIf 
\end{algorithmic}
\end{algorithm}
\vspace{-10pt}

\paragraph{Objective}
We optimize GA-GRPO under the GRPO objective without the KL term, directly maximizing grounding-aligned returns. We use a group-relative advantage modulated by the tool-grounding signal; the full objective and notation are provided in Appendix~\ref{appendix:GA-GRPO objective}.

\textbf{Trajectory-Level Fidelity-Weighted Advantage.}
To encourage grounded tool use, we modulate each trajectory advantage by tool-outcome fidelity. We set $f_i = R_{\text{evidence}}$ and let $m_i\in\{0,1\}$ indicate tool usage. We define:

\begin{equation}
\hat{A}_i = A_i \cdot h(\tau_i),
\end{equation}
where $h(\tau_i)$ keeps tool-free trajectories unchanged and adjusts tool-using ones by fidelity and optimization direction:
\begin{equation}
\tilde{f}_i = \mathrm{clip}(f_i,\ -c,\ c),
\end{equation}
\begin{equation}
h(\tau_i)=
\begin{cases}
\max(1+\alpha \tilde{f}_i,\, s_{\min}), & A_i \ge 0,\ m_i=1,\\
\max(1-\alpha \tilde{f}_i,\, s_{\min}), & A_i < 0,\ m_i=1,\\
1, & m_i=0,
\end{cases}
\end{equation}
with $\alpha>0$, clipping bound $c$, and floor $s_{\min}>0$. Intuitively, higher fidelity boosts positive advantages and softens negative ones, avoiding over-penalizing grounded attempts while still discouraging ungrounded tool use.

\section{Experiments}

\begin{table*}[t]
\centering
\scriptsize
\setlength{\tabcolsep}{4.5pt}
\renewcommand{\arraystretch}{1.0}

\caption{\textbf{Comparison of MedScope with zero-shot LMMs across multi-grained video understanding tasks on SVU-31K.} CI, DO, CU, and TU denote correctness of information, detail orientation, contextual understanding, and temporal understanding, respectively. Best results are in \textbf{bold} and second-best results are \underline{underlined}.}

\vspace{-5pt}
\begin{tabular*}{\textwidth}{@{\extracolsep{\fill}} l|c|c|cccc|cccc}
\toprule
\multirow{3}{*}{Model}  & \multirow{3}{*}{\makecell[c]{Tool\\Use}} & \multirow{3}{*}{Reasoning}
& \multicolumn{4}{c|}{Full Video Description}
& \multicolumn{4}{c}{Fine-grained Video Description} \\
\cmidrule(lr){4-7}\cmidrule(lr){8-11}
& & & CI & DO & CU & TU & CI & DO & CU & TU \\
\midrule
Video-LLaVA-7B      & \xmark & \xmark & 1.03 & 1.01 & 1.01 & 1.03 & 1.15 & 1.09 & 1.24 & 1.21 \\
GroundingGPT-7B     & \xmark & \xmark & 1.05 & 1.06 & 1.15 & 1.08 & 1.05 & 1.16 & 1.26 & 1.16 \\
VideoLLaMA3-7B      & \xmark & \xmark & 1.66 & 1.71 & 2.21 & 1.86 & 1.58 & 1.87 & 2.29 & 1.95 \\
Qwen2.5-VL-7B       & \xmark & \xmark & 1.22 & 1.09 & 1.64 & 1.53 & 1.82 & 1.77 & 2.49 & 1.95 \\
InternVL3-8B       & \xmark & \xmark & 1.34 & 1.11 & 1.52 & 1.63 & 1.76 & 1.67 & 2.31 & 2.02 \\
SurgVidLM           & \xmark & \xmark & 2.40 & 2.11 & 2.69 & 2.22 & 2.27 & 2.62 & 3.06 & 2.44 \\
LongVT-7B-RFT       & \cmark & \cmark & 3.96 & 4.10 & 3.90 & 3.87 & 3.77 & \underline{4.10} & 3.78 & 3.76 \\
VideoR1-7B          & \xmark & \cmark & 4.25 & 4.33 & 4.25 & 4.24 & \underline{3.80} & 4.01 & \underline{3.79} & \underline{3.77} \\
VideoChat-R1-7B     & \cmark & \cmark & 4.01 & 4.12 & 3.92 & 4.02 & 3.30 & 3.41 & 3.29 & 3.37 \\
VideoRFT-7B         & \cmark & \cmark & \underline{4.45} & \underline{4.52} & \underline{4.44} & \underline{4.43} & 3.46 & 3.61 & 3.50 & 3.53 \\
\midrule
MedScope            & \cmark & \cmark & \textbf{4.76} & \textbf{4.77} & \textbf{4.77} & \textbf{4.75} & \textbf{4.00} & \textbf{4.36} & \textbf{4.15} & \textbf{4.08} \\
\midrule\midrule

\multirow{3}{*}{Model}  & \multirow{3}{*}{\makecell[c]{Tool\\Use}} & \multirow{3}{*}{Reasoning}
& \multicolumn{4}{c|}{Fine-grained Temporal Visual Reasoning}
& \multicolumn{4}{c}{Fine-grained Perception Visual Reasoning} \\
\cmidrule(lr){4-7}\cmidrule(lr){8-11}
& & & BLEU-4 & CIDEr & METEOR & ROUGE-L & BLEU-4 & CIDEr & METEOR & ROUGE-L \\
\midrule
Video-LLaVA-7B      & \xmark & \xmark & 5.65 & 2.17 & 11.59 & 19.73 & 10.76 & 5.48 & 13.52 & 25.93 \\
GroundingGPT-7B     & \xmark & \xmark & 2.81 & 1.44 & 10.52 & 17.86 & 4.47  & 1.51 & 11.96 & 21.34 \\
VideoLLaMA3-7B      & \xmark & \xmark & 7.68 & 1.92 & 11.53 & 23.19 & 9.57  & 3.97 & 12.63 & 24.21 \\
Qwen2.5-VL-7B       & \xmark & \xmark & 4.41 & 1.55 & 13.09 & 16.27 & 3.19  & 2.18 & 15.04 & 25.55 \\
InternVL3-8B       & \xmark & \xmark & 3.83 & 1.13 & 11.28 & 18.32 & 2.93 & 1.89 & 12.20 & 26.72 \\
SurgVidLM           & \xmark & \xmark & \underline{10.10} & \underline{3.83} & 14.13 & \underline{31.58} & \underline{16.71} & \underline{9.76} & 18.27 & \underline{37.51} \\
LongVT-7B-RFT       & \cmark & \cmark  & 6.22 & 1.77 & \underline{14.99} & 21.26 & 5.41 & 2.00 & \underline{19.53} & 25.59 \\
VideoR1-7B          & \xmark & \cmark & 8.25 & 3.13 & 14.23 & 24.41 & 7.45 & 5.13 & 13.91 & 23.34 \\
VideoChat-R1-7B     & \cmark & \cmark & 4.30 & 1.56 & 11.89 & 19.83 & 5.90 & 1.93 & 17.62 & 24.54 \\
VideoRFT-7B         & \cmark & \cmark & 6.45 & 2.52 & 11.44 & 20.43 & 6.46 & 3.61 & 16.90 & 23.78 \\
\midrule
MedScope            & \cmark & \cmark & \textbf{11.30} & \textbf{4.56} & \textbf{16.89} & \textbf{34.83} & \textbf{18.90} & \textbf{10.93} & \textbf{21.62} & \textbf{41.54} \\
\bottomrule
\end{tabular*}

\label{tab:svu31k_main}
\vspace{-5pt}
\end{table*}

\begin{table*}[!h]
\centering
\setlength{\tabcolsep}{3.8pt}
\renewcommand{\arraystretch}{1.0}
\scriptsize

\caption{\textbf{Performance comparison on ClinVideo-Eval.} We report in-domain results on OphVL and MedVideoCap and out-of-domain generalization on SurgVidLM for temporal grounding and grounded VQA. Agent baselines can invoke tools, while other models do not.}

\vspace{-5pt}
\resizebox{\textwidth}{!}{%
\begin{tabular}{l|c|c|c|c|c|c|c|c|c|c|c|c|c|c}
\toprule
\multirow{4}{*}{\textbf{Model}}
& \multicolumn{8}{c}{\textbf{In-domain}}
& \multicolumn{6}{c}{\textbf{Out-of-domain}} \\
\cmidrule(lr){2-9}\cmidrule(lr){10-15}
& \multicolumn{6}{c}{\textbf{\shortstack[c]{OphVL\\(\(\approx\) 371147 sec)}}}
& \multicolumn{2}{c}{\textbf{\shortstack[c]{MedVideoCap\\(\(\approx\) 9575 sec)}}}
& \multicolumn{6}{c}{\textbf{\shortstack[c]{SurgVidLM\\(\(\approx\) 1320694 sec)}}} \\
\cmidrule(lr){2-7}\cmidrule(lr){8-9}\cmidrule(lr){10-15}
& \multicolumn{4}{c}{Temporal Grounding} & \multicolumn{2}{c}{Grounded VQA}
& \multicolumn{2}{c}{Grounded VQA}
& \multicolumn{4}{c}{Temporal Grounding} & \multicolumn{2}{c}{Grounded VQA} \\
\cmidrule(lr){2-5}\cmidrule(lr){6-7}
\cmidrule(lr){8-9}
\cmidrule(lr){10-13}\cmidrule(lr){14-15}
& R@0.3 & R@0.5 & R@0.7 & mIoU & mIoU & Acc
& mIoU & Acc
& R@0.3 & R@0.5 & R@0.7 & mIoU & mIoU & Acc \\
\midrule

\multicolumn{15}{c}{\textit{\textbf{Proprietary LMMs}}} \\
\midrule

\textcolor{gray}{GPT-4o}
& \textcolor{gray}{41.73} & \textcolor{gray}{34.33} & \textcolor{gray}{29.23} & \textcolor{gray}{35.75} & \textcolor{gray}{40.08} & \textcolor{gray}{9.79}
& \textcolor{gray}{88.80} & \textcolor{gray}{29.14}
& \textcolor{gray}{15.90} & \textcolor{gray}{9.00} & \textcolor{gray}{5.65} & \textcolor{gray}{13.35} & \textcolor{gray}{15.90} & \textcolor{gray}{6.85} \\

\textcolor{gray}{Gemini-2.5-Flash}
& \textcolor{gray}{59.33} & \textcolor{gray}{48.77} & \textcolor{gray}{13.50} & \textcolor{gray}{68.31} & \textcolor{gray}{43.38} & \textcolor{gray}{12.59}
& \textcolor{gray}{76.74} & \textcolor{gray}{26.96}
& \textcolor{gray}{43.51} & \textcolor{gray}{33.89} & \textcolor{gray}{23.64} & \textcolor{gray}{33.95} & \textcolor{gray}{12.94} & \textcolor{gray}{4.36} \\

\textcolor{gray}{Gemini 3-Pro-Preview}
& \textcolor{gray}{87.85} & \textcolor{gray}{73.06} & \textcolor{gray}{57.57} & \textcolor{gray}{72.43} & \textcolor{gray}{73.81} & \textcolor{gray}{34.12}
& \textcolor{gray}{96.46} & \textcolor{gray}{32.31}
& \textcolor{gray}{45.19} & \textcolor{gray}{31.59} & \textcolor{gray}{22.80} & \textcolor{gray}{37.35} & \textcolor{gray}{35.28} & \textcolor{gray}{21.99} \\

\midrule
\multicolumn{15}{c}{\textit{\textbf{Open-Source LMMs}}} \\
\midrule

Qwen2.5-VL-7B-Instruct
& 43.45 & 41.16 & 21.94 & 30.18 & 32.91 & 4.37
& 56.87 & 25.86
& 27.61 & 21.94 & 15.31 & 19.15 & 11.98 & 3.73 \\

InternVL3-8B
& 42.15 & 36.65 & 22.57 & 29.01 & 39.48 & 9.79
& \underline{57.62} & 26.61
& 20.92 & 14.23 & 10.88 & 18.55 & 16.6 & 3.53 \\

VideoLLaMA3-7B
& 37.36 & 25.01 & 14.39 & 23.10 & 20.78 & 1.34
& 42.14 & 15.89
& 12.67 & 10.08 & 5.21 & 9.37 & 8.55 & 1.93 \\

\midrule
\multicolumn{15}{c}{\textit{\textbf{Reasoning LMMs}}} \\
\midrule

Video-R1-7B
& 63.94 & 22.89 & 7.04 & 33.87 & 41.60 & \underline{27.45}
& 38.37 & \underline{34.55}
& 41.18 & 23.07 & 9.42 & 31.45 & 31.61 & 21.05 \\

VideoChat-R1-7B
& 30.58 & 21.41 & 9.17 & 30.93 & 32.40 & 14.68
& 23.13 & 33.69
& 46.87 & 26.04 & 3.15 & 21.43 & 29.29 & 15.97 \\

Video-RFT
& 26.41 & 19.37 & 7.04 & 27.06 & 28.7 & 15.24
& 31.08 & 28.47
& \underline{60.67} & 23.01 & 4.18 & 43.26 & 29.21 & 19.3 \\

\midrule
\multicolumn{15}{c}{\textit{\textbf{Reasoning Agents}}} \\
\midrule

LongVT-7B-RFT
& 48.73 & 32.32 & 23.52 & 39.56 & 42.17 & 14.16
& 55.34 & 31.22
& 51.13 & 37.66 & 16.73 & 39.74 & 25.72 & 17.02 \\

ReWatch-R1-7B
& 45.85 & 30.56 & 26.52 & 24.77 & 36.20 & 16.78
& 35.25 & 25.48
& \best{69.96} & \underline{39.75} & 10.46 & \best{47.17} & 25.47 & 21.99 \\

\rowcolor{gray!12}
MedScope-7B-SFT
& \underline{67.19} & \underline{54.43} & \underline{43.72} & \underline{49.28} & \underline{53.28} & 25.03
& 52.7 & 32.2
& 52.04 & 38.75 & \best{22.97} & 41.63 & \underline{32.13} & \underline{35.2} \\

\rowcolor{gray!12}
MedScope-7B-RL
& \best{81.92} & \best{68.87} & \best{54.35} & \best{64.42} & \best{65.58} & \best{35.10}
& \best{61.20} & \best{39.80}
& 55.60 & \best{40.10} & \underline{22.35} & \underline{44.30} & \best{34.10} & \best{37.90} \\

\rowcolor{gray!12}
$\Delta$ (vs Qwen2.5-VL-7B)
& \textcolor{red!85}{+38.5} & \textcolor{red!70}{+27.7} & \textcolor{red!85}{+32.4} & \textcolor{red!85}{+34.2} & \textcolor{red!85}{+32.7} & \textcolor{red!85}{+30.7}
& \textcolor{red!35}{+4.3} & \textcolor{red!55}{+13.9}
& \textcolor{red!70}{+28.0} & \textcolor{red!55}{+18.2} & \textcolor{red!35}{+7.0} & \textcolor{red!70}{+25.1} & \textcolor{red!70}{+22.1} & \textcolor{red!85}{+34.2} \\
\bottomrule
\end{tabular}%
}
\label{tab:medscope_indomain_outdomain}
\vspace{-12pt}
\end{table*}

\subsection{Experimental Setup}

\paragraph{Benchmarks}
\label{Benchmarks}
We evaluate our model on two medical video benchmarks covering multi-grained video understanding, visual reasoning, and grounded clinical question answering. 
The first benchmark, \textbf{SVU-31K} \citep{wang2025surgvidlm}, focuses on multi-grained video description and visual reasoning, including Full and Fine-grained Video Description as well as Fine-grained Temporal and Perception Visual Reasoning. 
The second benchmark, \textbf{ClinVideo-eval}, evaluates grounded medical video understanding under both in-domain and out-of-domain settings, spanning OphVL, MedVideoCap, and SurgVidLM. It supports Temporal Grounding and Grounded VQA tasks, enabling systematic evaluation of generalization from in-domain to out-of-domain scenarios. Additional details are provided in the Appendix~\ref{appendix:Benchmarks}.

\paragraph{Implementation Details.}
We take Qwen2.5-VL-7B-Instruct as the base model for all experiments. For SFT, we use \texttt{LMMs-Engine}~\citep{lmms_engine2025}, while RL is implemented with \texttt{verl}~\citep{sheng2024hybridflow} and rollouts are generated using \texttt{SGLang}~\citep{zheng2024sglang}. For cold-start SFT, we jointly train on \textbf{ClinVideo-CoT} and \textbf{ClinVideo-VCoT} for one epoch using 32 H200 GPUs. For RL, we train on \textbf{ClinVideo-RL} for one epoch using 64 H200 GPUs. Additional optimization settings and hyperparameters are provided in the Appendix~\ref{appendix:Implementation Details}.

\paragraph{Baseline Methods \& Evaluation Metrics}
\label{Baseline Methods & Evaluation Metric}
We compare MedScope with diverse baselines, including proprietary models, open-source video-language models, reasoning models, and tool-using agents. We follow standard medical video protocols to evaluate video description, visual reasoning, and grounded understanding in both in-domain and out-of-domain settings. Full details are provided in Appendix~\ref{appendix:Baselines & Metrics}.

\subsection{Main Results}

\begin{table*}[!t]
\centering
\scriptsize
\setlength{\tabcolsep}{3.6pt}
\renewcommand{\arraystretch}{1.0}

\caption{\textbf{Ablations on training stages and VCoT supervision.}
\textbf{Top: Training stages.} We ablate Warm-up, SFT, and RL and evaluate combinations on SVU-31K.
\textbf{Bottom: VCoT supervision.} We remove VCoT trajectories during SFT and compare SFT-only and SFT+RL.}
\vspace{-5pt}

\begin{tabular*}{\textwidth}{@{\extracolsep{\fill}} l|ccc|ccccc|ccccc}
\toprule
\multirow{2}{*}{Setting} & \multicolumn{3}{c|}{Training Components} 
& \multicolumn{5}{c|}{Fine-grained Temporal Visual Reasoning}
& \multicolumn{5}{c}{Fine-grained Perception Visual Reasoning} \\
\cmidrule(lr){2-4}\cmidrule(lr){5-9}\cmidrule(lr){10-14}
& Warm-up & SFT & RL
& BLEU-4 & CIDEr & METEOR & ROUGE-L & Tool
& BLEU-4 & CIDEr & METEOR & ROUGE-L & Tool \\
\midrule

\multicolumn{14}{c}{\textit{\textbf{Training Components}}} \\
\midrule
\rowcolor{gray!50}
Qwen2.5-VL-7B-Instruct
& \xmark & \xmark & \xmark
& 4.41 & 1.55 & 13.09 & 16.27 & 7.8\%
& 3.19 & 2.18 & 15.04 & 25.55 & 6.4\% \\

Warm-up only
& \cmark & \xmark & \xmark
& 5.13 & 2.16 & 12.98 & 17.36 & 8.6\%
& 7.01 & 4.54 & 16.40 & 26.72 & 6.9\% \\

Warm-up + SFT
& \cmark & \cmark & \xmark
& 7.64 & 2.59 & 14.51 & 23.43 & 46.2\%
& 11.76 & 5.21 & 17.49 & 35.90 & 34.7\% \\

Warm-up + RL
& \cmark & \xmark & \cmark
& \underline{7.70} & \underline{4.70} & \textbf{17.05} & \underline{24.50} & 27.4\%
& \underline{12.10} & \underline{5.40} & \underline{17.70} & \underline{36.20} & 20.7\% \\

\rowcolor{gray!12}
Warm-up + SFT + RL (MedScope)
& \cmark & \cmark & \cmark
& \textbf{11.30} & \textbf{4.56} & \underline{16.89} & \textbf{34.83} & 63.8\%
& \textbf{18.90} & \textbf{10.93} & \textbf{21.62} & \textbf{41.54} & 51.8\% \\

\midrule

\multicolumn{14}{c}{\textit{\textbf{Data Recipe: Effect of VCoT}}} \\
\midrule
Warm-up + SFT (w/o VCoT)
& \cmark & \cmark & \xmark
& 6.92 & 2.31 & 14.06 & 22.35 & 8.9\%
& 10.88 & 4.93 & 17.12 & 34.70 & 11.0\% \\

Warm-up + SFT (w/ VCoT)
& \cmark & \cmark & \xmark
& 7.64 & 2.59 & 14.51 & 23.43 & 46.2\%
& 11.76 & 5.21 & 17.49 & 35.90 & 34.7\% \\

Warm-up + SFT + RL (w/o VCoT)
& \cmark & \cmark & \cmark
& \underline{7.95} & \underline{4.13} & \underline{15.24} & \underline{25.40} & 25.5\%
& \underline{12.60} & \underline{5.80} & \underline{18.05} & \underline{37.10} & 21.3\% \\

\rowcolor{gray!12}
Warm-up + SFT + RL (w/ VCoT)
& \cmark & \cmark & \cmark
& \textbf{11.30} & \textbf{4.56} & \textbf{16.89} & \textbf{34.83} & 63.8\%
& \textbf{18.90} & \textbf{10.93} & \textbf{21.62} & \textbf{41.54} & 51.8\% \\

\bottomrule
\end{tabular*}

\label{tab:ab_train_and_vcot_svu_reasoning}
\vspace{-15pt}
\end{table*}

Table~\ref{tab:svu31k_main} reports results on SVU-31K across multi-grained video description and fine-grained temporal and perceptual visual reasoning. MedScope ranks first on both full-video and fine-grained description, achieving 4.77 DO on full-video description and 4.36 DO on fine-grained description, and consistently surpassing strong reasoning baselines. It also outperforms tool-using agents such as LongVT-7B-RFT as well as prior reasoning models including VideoRFT-7B and VideoR1-7B, showing that tool use is most effective when paired with verification-driven training. On fine-grained reasoning, MedScope sets a new state of the art, reaching 4.56 CIDEr on temporal reasoning and 10.93 CIDEr on perceptual reasoning, with clear gains over the medical baseline SurgVidLM. Overall, MedScope delivers robust improvements from global narrative fidelity to localized evidence reasoning on SVU-31K.

Table~\ref{tab:medscope_indomain_outdomain} reports results on the ClinVideo-Eval benchmark for temporal grounding and grounded VQA under in-domain and out-of-domain settings. MedScope-7B-RL achieves the strongest open-source performance overall, surpassing both reasoning models and tool-using agents. It reaches 64.42 mIoU for temporal grounding and 65.58 mIoU for grounded VQA on OphVL, and attains 61.20 mIoU and 39.80 accuracy on MedVideoCap, which contains shorter videos. On SurgVidLM, MedScope remains robust out of domain and competitive with closed-source models, while consistently improving over open-source baselines.

\subsection{Ablation Studies}

\noindent\textbf{Warm-up supplies basic clinical reasoning, SFT learns tool use, and RL optimizes verification decisions.}
Table~\ref{tab:ab_train_and_vcot_svu_reasoning} (top) shows that Warm-up alone brings limited gains, while removing SFT severely degrades tool-conditioned verification, indicating that RL by itself is insufficient to induce reliable evidence seeking and often converges to shallow, instruction-following behaviors with fewer tool calls. Warm-up+SFT yields consistent improvements by imitating correct tool trajectories, but remains more vulnerable to distribution shift without decision-level refinement. Adding RL on top of SFT further strengthens evidence-grounded decision making, consolidating when to retrieve additional evidence and when to commit, and delivers the best overall results across both temporal and perceptual reasoning.

\noindent\textbf{High-quality VCoT cold-start trajectories are critical for effective RL.}
Table~\ref{tab:ab_train_and_vcot_svu_reasoning} (bottom) shows that removing VCoT during SFT consistently hurts downstream reasoning. Without VCoT, the model learns weaker tool-conditioned evidence aggregation and fine-grained perception, leading to uniformly lower scores. The gap further widens after RL: starting RL from VCoT-free SFT yields only limited gains, suggesting unstable tool invocation and poorer credit assignment. In contrast, VCoT-enabled cold-start provides a reliable initialization for RL, producing the best overall performance across both temporal and perceptual reasoning.

\begin{figure}[t]
    \centering
    \includegraphics[width=0.95\columnwidth]{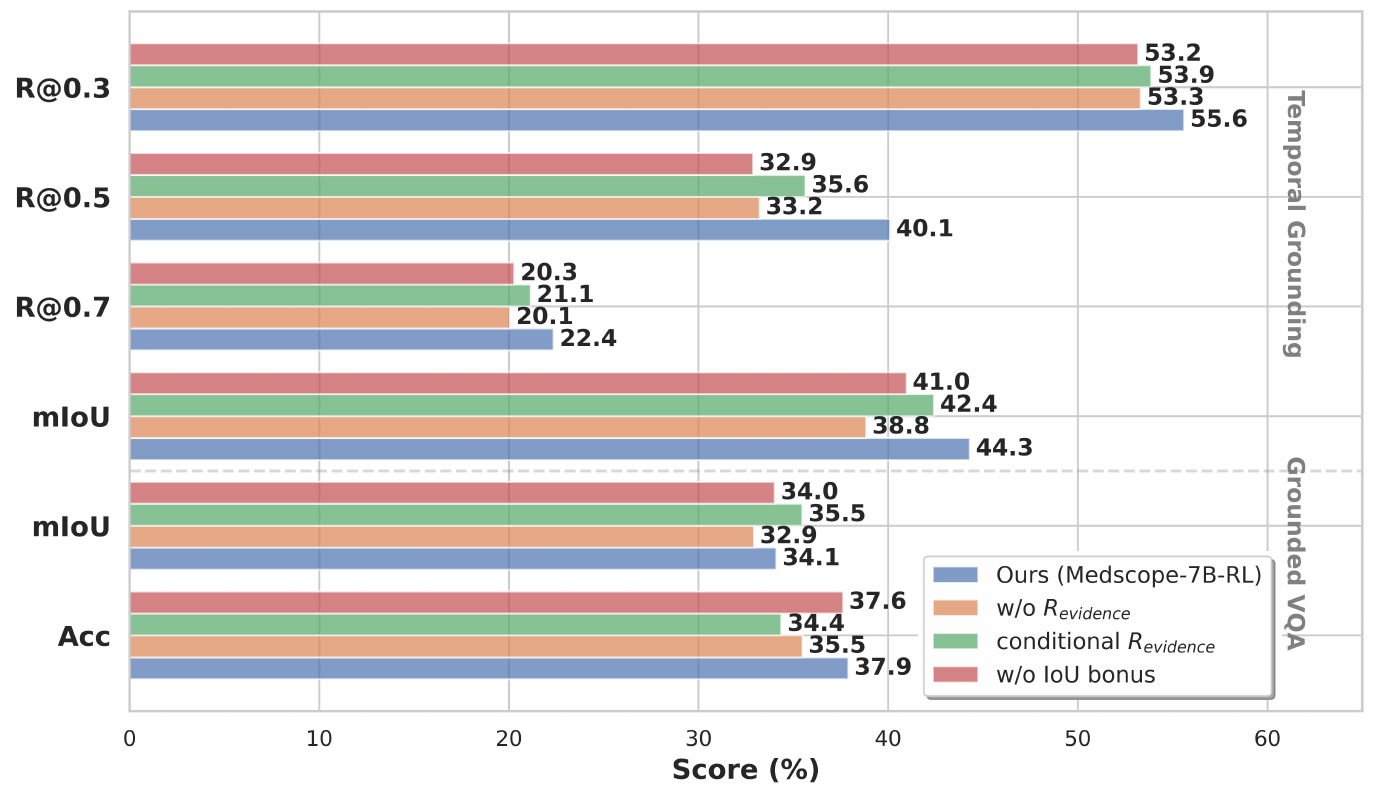}
    \vspace{-3pt}
    \caption{
\textbf{Ablations on reward design.} We compare \textbf{Ours} with three variants: \textbf{w/o $R_{\text{evidence}}$} removes the evidence reward; \textbf{conditional $R_{\text{evidence}}$} applies it only when $R_{\text{acc}}{=}1$; \textbf{w/o IoU bonus} removes the continuous IoU bonus in $R_{\text{evidence}}$.
}

    \vspace{-12pt}
    \label{fig:fig5}
\end{figure}

\noindent\textbf{Grounding-aware reward design is essential for evidence-faithful temporal reasoning.}
Figure~\ref{fig:fig5} reports reward ablations on temporal grounding and grounded VQA. 
Removing the evidence reward lowers localization quality substantially, with R@0.5 dropping from 40.1 to 33.2 and mIoU from 44.3 to 38.8, showing that answer-only supervision does not provide enough signal to learn reliable evidence selection. 
Conditioning the evidence reward on answer correctness improves over this variant, but still trails the full design, suggesting that decoupled grounding feedback is important even when intermediate attempts are imperfect. 
Removing the IoU bonus further weakens precise alignment, most clearly at stricter criteria where R@0.7 decreases from 22.4 to 20.3, highlighting the benefit of continuous, overlap-sensitive feedback for tightening temporal localization.



\section{Conclusion}
We present \textbf{MedScope}, a tool-using clinical LMM that enables ``think with videos'' via coarse-to-fine temporal evidence seeking and frame-level verification in long-form procedures. We introduce \textbf{ClinVideoSuite}, a large-scale suite with localized supervision, grounded QA, and environment-interactive visual-CoT trajectories, and propose \textbf{GA-GRPO} to reinforce temporally aligned tool use with grounding-aware rewards and evidence-modulated advantages. Experiments on SVU-31K and ClinVideo-Eval show state-of-the-art results on multi-grained video understanding, fine-grained reasoning, and grounded VQA across diverse benchmarks. We hope this work encourages more reliable verification and evidence attribution for clinical video intelligence.



\section*{Impact Statement}
This work aims to advance evidence-grounded clinical video understanding and reasoning for research purposes. All clinical data used to construct our benchmarks and training suite were obtained under institutional ethical review and approval, and were processed in accordance with applicable regulations and data-governance requirements, including privacy protection and de-identification where required. While our approach has potential benefits for clinical education and decision support, it may also introduce risks if deployed improperly, such as over-reliance on automated outputs or failure under distribution shifts. We therefore emphasize that MedScope is not intended for standalone clinical use, and we encourage future work on rigorous prospective validation, robustness evaluation, and responsible deployment practices with appropriate human oversight.

\nocite{langley00}

\bibliography{example_paper}
\bibliographystyle{icml2026}

\newpage
\appendix
\onecolumn

\section{Dataset Statistics}
\label{appendix:Dataset Statistics}

\begin{figure*}[ht]
    \centering
    \includegraphics[width=0.92\linewidth]{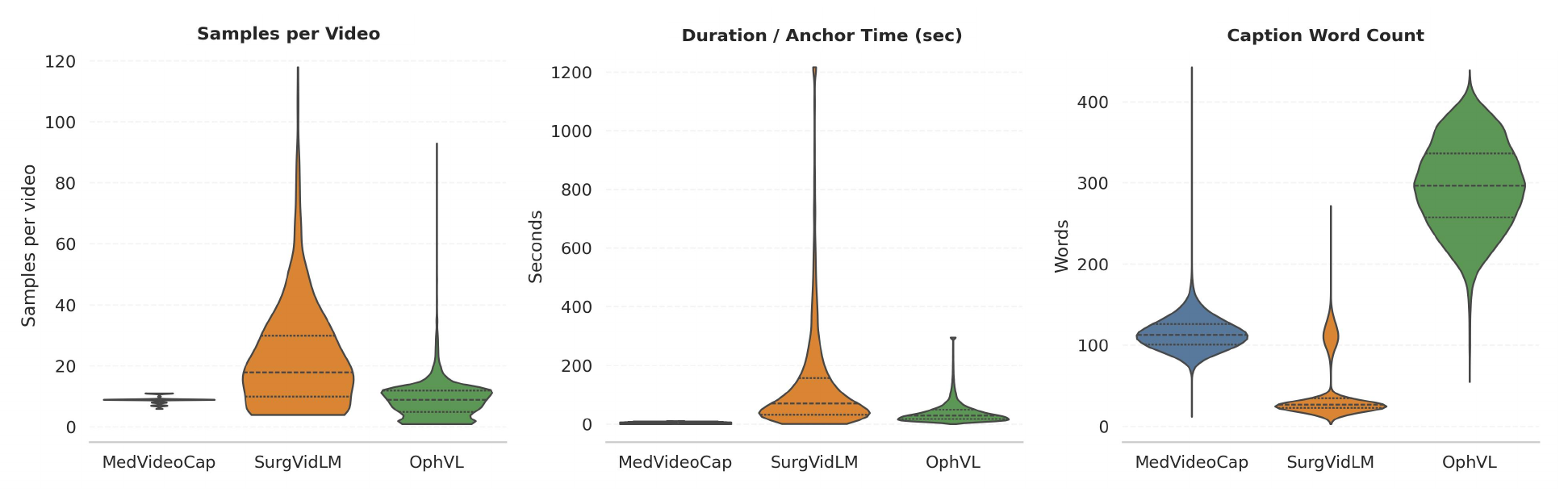}
    \caption{Overview of data characteristic distributions across different data sources used in constructing ClinVideo-Cap-635K.}
    \label{fig:data1}
\end{figure*}

\begin{figure*}[ht]
    \centering
    \includegraphics[width=0.92\linewidth]{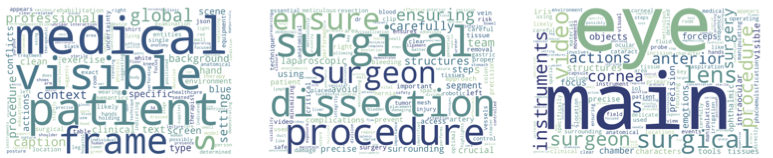}
    \caption{Word Clouds of ClinVideo-Cap-635K (from left to right: MedVideoCap, SurgVidLM, and OphVL).}
    \label{fig:data1-2}
\end{figure*}

\paragraph{\textbf{ClinVideo-Cap-635K.}}
ClinVideo-Cap-635K contains 634.8K timestamped dense captions sourced from MedVideoCap~\citep{wang2025medgen}, OphVL~\citep{hu2025ophclip}, and SurgVidLM~\citep{wang2025surgvidlm}. Figure~\ref{fig:data1} summarizes the key distributional characteristics across sources. SurgVidLM provides the densest temporal supervision, with an average of approximately 22 captions per video and a long-tailed duration distribution whose 95th percentile extends to several hundred seconds, reflecting the prolonged and continuous nature of surgical workflows. In contrast, MedVideoCap videos are shorter and more uniform, typically containing fewer than 9 captions per video with tightly distributed anchor times. OphVL exhibits moderate caption density but substantially longer captions, with average word counts often exceeding 185 words. These linguistic characteristics are further illustrated in Figure~\ref{fig:data1-2}.

\begin{figure*}[!h]
    \centering
    \includegraphics[width=0.92\linewidth]{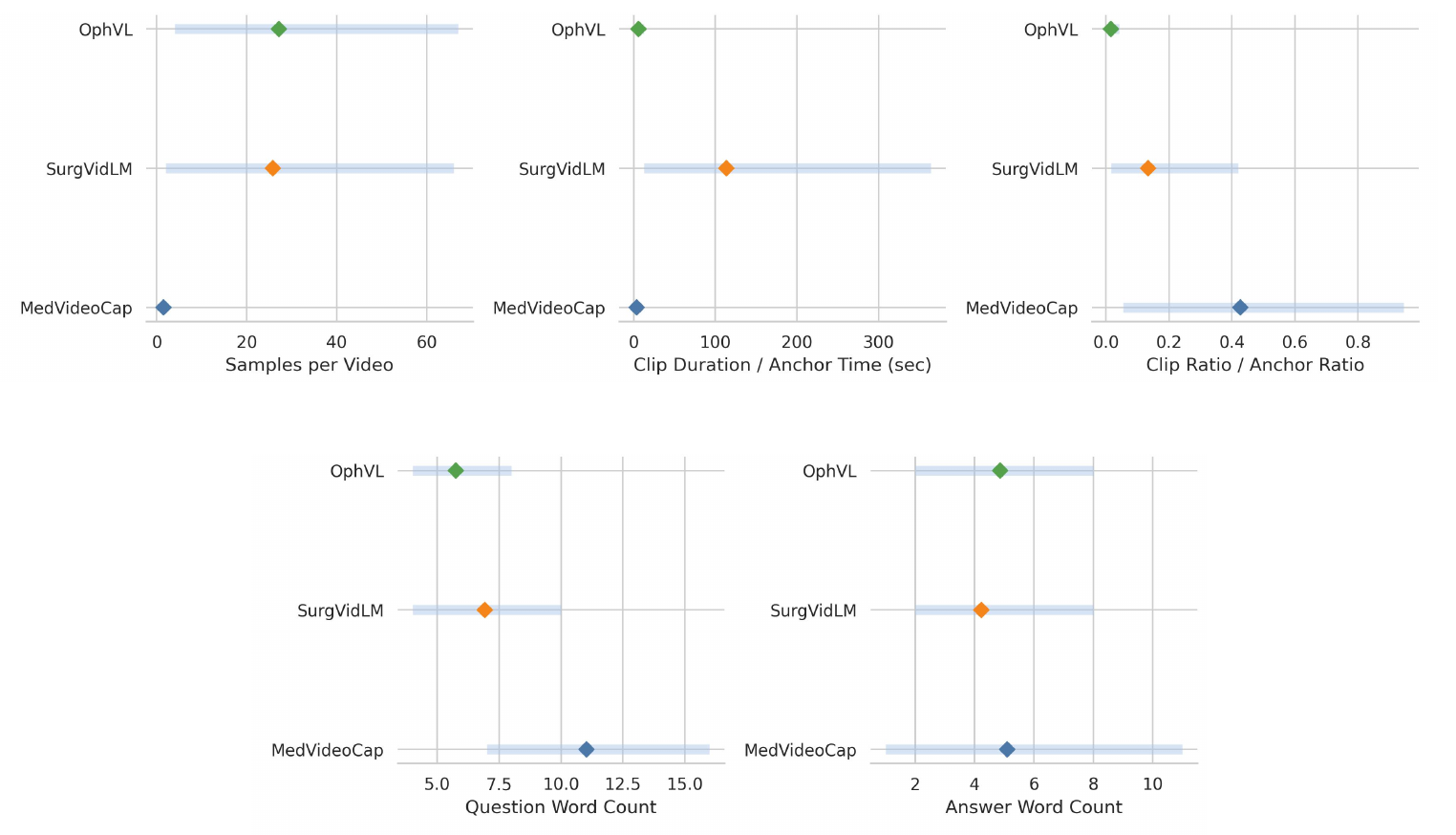}
    \caption{Overview of data characteristic distributions across different data sources used in constructing ClinVideo-QA-254K. Diamond markers indicate the mean, and the light-blue bars represent the 5th–95th percentile range.}
    \label{fig:data2}
\end{figure*}

\begin{figure*}[!h]
    \centering
    \includegraphics[width=0.92\linewidth]{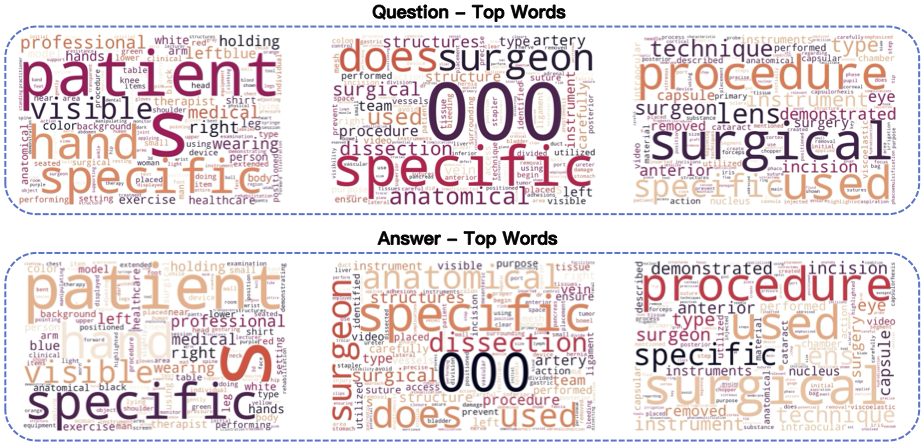}
    \caption{Word Clouds of ClinVideo-QA-254K for Questions and Answers (from left to right: MedVideoCap, SurgVidLM, and OphVL).}
    \label{fig:data2-2}
\end{figure*}

\paragraph{\textbf{ClinVideo-QA-254K.}}
ClinVideo-QA-254K consists of 253.8K open-ended and evidence-grounded question answering pairs with localized supervision windows. As shown in Figure~\ref{fig:data2}, questions remain concise across all sources, with average lengths of roughly 6 to 11 words, while answers are short and precise at about 4 to 5 words. We also observe source-specific differences in evidence granularity: SurgVidLM QA pairs are associated with longer clip durations and higher clip ratios, whereas OphVL questions more frequently rely on shorter and more localized visual evidence. Complementary word cloud visualizations for questions and answers are provided in Figure~\ref{fig:data2-2}.

\paragraph{\textbf{Professional categorization.}}
To characterize clinical coverage beyond basic statistics, we perform a fine-grained professional categorization of ClinVideo-QA-254K using a two-level hierarchy, where each item is jointly interpreted from its question and answer by \texttt{gemini-2.5-flash}~\citep{comanici2025gemini}. Figure~\ref{fig:data3} reports the hierarchical category distribution for MedVideoCap, while Figures~\ref{fig:data3-2} and~\ref{fig:data3-3} present the corresponding distributions for SurgVidLM and OphVL, respectively. At the top level, the dataset covers 21 distinct clinical areas across the three sources, and dominant categories closely align with real-world practice. Clinical Practice accounts for approximately 40\% of MedVideoCap QA pairs, while procedure-centric topics dominate surgical domains, including Lens and Vitreous Procedures in OphVL at about 40\% and Robotic Tissue Manipulation in SurgVidLM at about 30\%. At the second level, the majority of QA pairs concentrate on fine-grained procedural reasoning, such as treatment and operative steps accounting for over 60\% within Clinical Practice, lens fragmentation and removal at about 59\% within OphVL, and robotic tissue dissection at about 70\% within SurgVidLM. These results indicate that \textbf{ClinVideo-QA-254K} spans broad clinical coverage while placing strong emphasis on action-centric and temporally grounded visual understanding.


\section{Details of Algorithm}

\subsection{Warm-Up SFT Objective}
\label{appendix:warmup_sft_objective}

Let $\mathcal{D}_{\mathrm{vid\text{-}cot}}$, $\mathcal{D}_{\mathrm{long\text{-}cot}}$, and $\mathcal{D}_{\mathrm{img\text{-}vqa}}$ denote the three warm-up corpora (Video-R1 CoT, LongVideo-Reason subset, and surgical image-level VQA, respectively). We form a single merged training set
$\mathcal{D}_{\mathrm{warm}}=\mathcal{D}_{\mathrm{vid\text{-}cot}}\cup \mathcal{D}_{\mathrm{long\text{-}cot}}\cup \mathcal{D}_{\mathrm{img\text{-}vqa}}$
by concatenating all examples and shuffling, while preserving each source's supervision format.
Each example is represented as a token sequence $y=(y_1,\ldots,y_{|y|})$ conditioned on context $x$ (prompts and multimodal inputs). We train the policy $\pi_\theta$ with the standard teacher-forcing objective, minimizing the token-level negative log-likelihood over the merged corpus:
\begin{equation}
\mathcal{L}_{\mathrm{warm}}(\theta)
=
\mathbb{E}_{(x,y)\sim\mathcal{D}_{\mathrm{warm}}}
\left[
-\sum_{t=1}^{|y|}
\log \pi_\theta\!\left(y_t \mid x, y_{<t}\right)
\right].
\label{eq:warmup_sft_nll}
\end{equation}
We implement this objective by sampling mini-batches from $\mathcal{D}_{\mathrm{warm}}$ after global shuffling, ensuring that the model is jointly exposed to video CoT, long-video CoT, and image-level VQA signals during warm-up.

\subsection{GA-GRPO Reward Details}
\label{app:reward_details}

\paragraph{Composite reward.}
For each rollout trajectory $\tau$ with terminal prediction $y$ and tool-call trace $T$, we define a dense composite reward
\begin{equation}
R(\tau)=R_{\text{acc}}(y)+R_{\text{format}}(\tau)+R_{\text{evidence}}(T).
\end{equation}
Here $R_{\text{acc}}$ measures answer correctness, $R_{\text{format}}$ enforces the structured interface, and $R_{\text{evidence}}$ evaluates tool-grounding quality based on the last tool call.

\paragraph{Answer reward $R_{\text{acc}}$.}
We use an LLM-as-a-judge to score the final answer with a three-level grading scheme to reduce reward sparsity:
\begin{equation}
R_{\text{acc}}(y)=
\begin{cases}
1, & \text{correct},\\
0.5, & \text{partially correct},\\
0, & \text{incorrect or overlong}.
\end{cases}
\end{equation}

\paragraph{Format reward $R_{\text{format}}$.}
We apply a deterministic validator for the required trajectory interface. The format reward is binary:
\begin{equation}
R_{\text{format}}(\tau)=\mathbf{1}\bigl[\textsc{FormatOK}(\tau)\bigr],
\end{equation}
where \textsc{FormatOK} checks the presence and ordering constraints of \texttt{<think>}, \texttt{<tool\_call>} (optional), \texttt{<tool\_response>} (if a tool is called), and \texttt{<answer>}.

\paragraph{Tool-grounding reward $R_{\text{evidence}}$.}
We compute $R_{\text{evidence}}$ from the last tool call in $T$, aligning the reward with the native tool interface. If no tool is invoked, we set $R_{\text{evidence}}=0$. If the last tool call is \texttt{crop\_video}, we score interval alignment; if it is \texttt{get\_frame}, we score timestamp alignment. When both tools appear in a rollout, we still use the last call to define the primary grounding signal (consistent with the execution trace used during rollout).

\paragraph{\texttt{crop\_video} reward.}
Let the last \texttt{crop\_video} call predict an interval $[p_s,p_e]$. When interval supervision $[g_s,g_e]$ is available, we compute the temporal intersection-over-union
$\mathrm{IoU}([p_s,p_e],[g_s,g_e])=\frac{|[p_s,p_e]\cap[g_s,g_e]|}{|[p_s,p_e]\cup[g_s,g_e]|}$
and map it to a monotonic bonus schedule
$
R_{\text{crop}}=
\begin{cases}
\alpha\cdot \mathrm{sign}(\mathrm{IoU}-h_0)+\eta\left\lfloor\frac{\mathrm{IoU}-h_0}{\Delta}\right\rfloor, & \mathrm{IoU}>0,\\
0, & \mathrm{IoU}=0,
\end{cases}
$
where $h_0$ is a base overlap threshold, $\Delta$ is the step size, and $\alpha,\eta$ control the base and incremental bonus magnitudes. When only frame supervision $g_f$ is provided, we instead use a coverage indicator $R_{\text{crop}}=\mathbf{1}[p_s \le g_f \le p_e]$.

\paragraph{\texttt{get\_frame}: interval or frame supervision.}
Let the last \texttt{get\_frame} call query timestamp $t$. Under interval supervision $[g_s,g_e]$, we use an in-segment indicator:
\begin{equation}
R_{\text{frame}}=
\mathbf{1}\bigl[g_s \le t \le g_e\bigr].
\end{equation}
Under frame supervision $g_f$, we use a tolerance-based proximity score:
\begin{equation}
R_{\text{frame}}=
\max\!\left(0,\,1-\frac{|t-g_f|}{w}\right),
\end{equation}
where $w$ is a tolerance window controlling how quickly the reward decays as the query moves away from $g_f$.

\paragraph{Final grounding term.}
The tool-grounding reward sums the interval- and timestamp-level components:
\begin{equation}
R_{\text{evidence}}(T)=
\mathbf{1}\!\left[\exists\,\texttt{crop\_video}\in T\right]\cdot R_{\text{crop}}
\;+\;
\mathbf{1}\!\left[\exists\,\texttt{get\_frame}\in T\right]\cdot R_{\text{frame}},
\end{equation}
where $R_{\text{crop}}$ is computed from the last \texttt{crop\_video} call (if any) and $R_{\text{frame}}$ from the last \texttt{get\_frame} call (if any). If no tool call is made, both indicators are zero and $R_{\text{evidence}}(T)=0$. We use this grounding term together with $R_{\text{acc}}$ and $R_{\text{format}}$ to encourage rollouts that are both correct and temporally grounded by tool-based visual rationalization.

\paragraph{GA-GRPO Objective}
\label{appendix:GA-GRPO objective}
We follow the official GRPO objective and remove the KL term, directly optimizing grounding-aligned returns. Following the variance-reduction intuition of DAPO~\citep{yu2025dapo}, we adopt a group-relative, evidence-weighted advantage and omit variance normalization. Let $G(q)=\{\tau_k\}_{k=1}^G$ denote the group of sampled trajectories for a query $q$. The objective is
\begin{equation}
\mathcal{J}(\theta)=
\mathbb{E}_{\substack{q\sim P(Q)\\ \{\tau_k\}\sim\pi_{\theta_{\mathrm{old}}}(\cdot\mid q)}}
\!\left[\frac{1}{G}\sum_{k=1}^{G}
\frac{\pi_{\theta}(\tau_k\mid q)}{\pi_{\theta_{\mathrm{old}}}(\tau_k\mid q)}\hat{A}_k\right].
\end{equation}

\subsection{GA-GRPO Group-Level Advantage.}
\label{appendix:GA-GRPO details}
Following the variance-reduction intuition of DAPO~\citep{yu2025dapo}, we compute a group-relative advantage to reduce instability from overly easy queries with saturated rewards and overly hard queries with uniformly low rewards. Concretely, we summarize token-level rewards into a trajectory score and mean-center it within the response group $G$ for the same query:
\begin{equation}
S_i=\sum_{t=1}^{T_i} r_{i,t},
\qquad
A_i=S_i-\frac{1}{|G|}\sum_{j\in G} S_j.
\end{equation}

\paragraph{Why Advantage Modulation is Necessary?}
Group-centering stabilizes GA-GRPO by removing query-specific bias, but it does not solve the fundamental credit-assignment issue: a trajectory-level advantage $A_i$ is a single scalar applied to all actions in the response. As a result, all steps are rewarded or penalized equally, regardless of whether a particular visual rationale is faithful or spurious. In tool-augmented reasoning, this is especially problematic: a trajectory can achieve high reward due to a correct final answer while still using incorrect visual evidence, or fail due to language errors despite correctly localized evidence. Without modulation, the policy is incentivized to reproduce any visual rationale that co-occurs with success, even if it is misaligned with the evidence.

\paragraph{Evidence-Aware Modulation as Targeted Credit Assignment.}
We therefore modulate the trajectory advantage by an evidence-quality factor computed from the same grounding signals used in the reward (crop IoU and frame alignment). This transforms the update from a uniform scalar into a selectively scaled signal that reflects whether visual evidence is reliable. In advantageous trajectories, high-fidelity evidence receives amplified credit while low-fidelity evidence is down-weighted; in disadvantageous trajectories, the opposite weighting increases blame for poor evidence while protecting good evidence from being over-penalized. This mechanism preserves the variance-reduction benefits of group-level baselines while directly addressing the core mismatch between trajectory-level rewards and step-level visual correctness, thereby aligning policy updates with grounded visual reasoning.

\section{Additional Evaluation Results}

\begin{table*}[!t]
\centering
\scriptsize
\setlength{\tabcolsep}{4.5pt}
\renewcommand{\arraystretch}{1.15}

\caption{\textbf{Toolbox-enabled comparison across multi-grained video understanding tasks on SVU-31K.}
All models are allowed to call the same native tools (\texttt{crop\_video}, \texttt{get\_frame}) during inference. CI, DO, CU, and TU denote correctness of information, detail orientation, contextual understanding, and temporal understanding, respectively. Best results are in \textbf{bold} and second-best results are \underline{underlined}.}

\begin{tabular*}{\textwidth}{@{\extracolsep{\fill}} l|c|c|cccc|cccc}
\toprule
\multirow{3}{*}{Model}  & \multirow{3}{*}{\makecell[c]{Tool\\Use}} & \multirow{3}{*}{Reasoning}
& \multicolumn{4}{c|}{Full Video Description}
& \multicolumn{4}{c}{Fine-grained Video Description} \\
\cmidrule(lr){4-7}\cmidrule(lr){8-11}
& & & CI & DO & CU & TU & CI & DO & CU & TU \\
\midrule
Video-LLaVA-7B      & \cmark & \xmark & 1.05 & 1.00 & 1.03 & 1.02 & 1.14 & 1.11 & 1.23 & 1.22 \\
GroundingGPT-7B     & \cmark & \xmark & 1.06 & 1.07 & 1.13 & 1.09 & 1.04 & 1.17 & 1.25 & 1.15 \\
VideoLLaMA3-7B      & \cmark & \xmark & 1.67 & 1.69 & 2.24 & 1.84 & 1.60 & 1.85 & 2.31 & 1.93 \\
Qwen2.5-VL-7B       & \cmark & \xmark & 1.24 & 1.10 & 1.62 & 1.55 & 1.80 & 1.79 & 2.47 & 1.97 \\
InternVL3-8B        & \cmark & \xmark & 1.33 & 1.13 & 1.54 & 1.61 & 1.77 & 1.66 & 2.30 & 2.04 \\
SurgVidLM           & \cmark & \xmark & 2.38 & 2.13 & 2.70 & 2.21 & 2.26 & 2.60 & 3.08 & 2.43 \\
LongVT-7B-RFT       & \cmark & \cmark & 3.95 & \underline{4.08} & 3.92 & 3.86 & 3.78 & 4.09 & 3.77 & 3.77 \\
VideoR1-7B          & \cmark & \cmark & 4.24 & 4.35 & 4.23 & 4.25 & \underline{3.79} & 4.03 & \underline{3.80} & \underline{3.76} \\
VideoChat-R1-7B     & \cmark & \cmark & 4.03 & 4.10 & 3.93 & 4.00 & 3.31 & 3.39 & 3.30 & 3.38 \\
VideoRFT-7B         & \cmark & \cmark & \underline{4.44} & \underline{4.53} & \underline{4.43} & \underline{4.41} & 3.47 & 3.60 & 3.52 & 3.52 \\
\midrule
MedScope            & \cmark & \cmark & \textbf{4.76} & \textbf{4.77} & \textbf{4.77} & \textbf{4.75} & \textbf{4.00} & \textbf{4.36} & \textbf{4.15} & \textbf{4.08} \\
\midrule\midrule

\multirow{3}{*}{Model}  & \multirow{3}{*}{\makecell[c]{Tool\\Use}} & \multirow{3}{*}{Reasoning}
& \multicolumn{4}{c|}{Fine-grained Temporal Visual Reasoning}
& \multicolumn{4}{c}{Fine-grained Perception Visual Reasoning} \\
\cmidrule(lr){4-7}\cmidrule(lr){8-11}
& & & BLEU-4 & CIDEr & METEOR & ROUGE-L & BLEU-4 & CIDEr & METEOR & ROUGE-L \\
\midrule
Video-LLaVA-7B      & \cmark & \xmark & 5.72 & 2.12 & 11.62 & 19.60 & 10.80 & 5.45 & 13.58 & 25.85 \\
GroundingGPT-7B     & \cmark & \xmark & 2.78 & 1.48 & 10.50 & 18.05 & 4.52  & 1.47 & 11.92 & 21.40 \\
VideoLLaMA3-7B      & \cmark & \xmark & 7.60 & 1.95 & 11.60 & 23.05 & 9.62  & 3.90 & 12.70 & 24.30 \\
Qwen2.5-VL-7B       & \cmark & \xmark & 4.48 & 1.52 & 13.15 & 16.40 & 3.16  & 2.20 & 14.98 & 25.70 \\
InternVL3-8B        & \cmark & \xmark & 3.80 & 1.18 & 11.20 & 18.45 & 2.98 & 1.85 & 12.30 & 26.60 \\
SurgVidLM           & \cmark & \xmark & \underline{10.05} & \underline{3.90} & 14.05 & \underline{31.70} & \underline{16.75} & \underline{9.70} & 18.35 & \underline{37.40} \\
LongVT-7B-RFT       & \cmark & \cmark  & 6.30 & 1.72 & \underline{14.95} & 21.10 & 5.45 & 1.98 & \underline{19.55} & 25.45 \\
VideoR1-7B          & \cmark & \cmark & 8.20 & 3.18 & 14.30 & 24.20 & 7.50 & 5.05 & 13.88 & 23.50 \\
VideoChat-R1-7B     & \cmark & \cmark & 4.25 & 1.60 & 11.85 & 20.00 & 5.95 & 1.90 & 17.70 & 24.40 \\
VideoRFT-7B         & \cmark & \cmark & 6.52 & 2.48 & 11.38 & 20.60 & 6.40 & 3.65 & 16.85 & 23.90 \\
\midrule
MedScope            & \cmark & \cmark & \textbf{11.30} & \textbf{4.56} & \textbf{16.89} & \textbf{34.83} & \textbf{18.90} & \textbf{10.93} & \textbf{21.62} & \textbf{41.54} \\
\bottomrule
\end{tabular*}

\label{tab:svu31k_main_toolbox}
\vspace{-5pt}
\end{table*}

In this section, we provide additional results to further contextualize our main findings. 
First, we report a toolbox-enabled comparison on SVU-31K under a unified inference interface, where all baselines are permitted to use the same native tools (\texttt{crop\_video}, \texttt{get\_frame}) to enable a more controlled assessment of tool access versus tool-conditioned reasoning (Table~\ref{tab:svu31k_main_toolbox}). 
Second, we present extra ablations on fidelity-weighted advantage modulation, isolating how different modulation designs affect the stability and effectiveness of grounded policy optimization (Figure~\ref{fig:fig6}).

\subsection{Toolbox-enabled comparison under a unified inference interface.}
To ensure a fairer comparison, we additionally evaluate all baselines under the same toolbox-enabled setting, where every LMM is allowed to invoke the native \texttt{crop\_video} and \texttt{get\_frame} tools at test time in Table~\ref{tab:svu31k_main_toolbox}.
Under this unified interface, most zero-shot LMMs show only small changes in scores, indicating that simply granting tool access is insufficient: these models typically do not learn when to call tools, which evidence window to retrieve, or how to integrate retrieved observations into a verifiable decision.
In contrast, \textsc{MedScope} consistently achieves the best performance across all task families, obtaining 4.76--4.77 on full-video description and 4.00/4.36/4.15/4.08 on fine-grained description (CI/DO/CU/TU), and outperforming strong reasoning baselines such as VideoRFT and VideoR1 even when they can use the same tools.
The gap is larger on fine-grained visual reasoning, where \textsc{MedScope} reaches 11.30 BLEU-4 and 4.56 CIDEr for temporal reasoning and 18.90 BLEU-4 and 10.93 CIDEr for perception reasoning, demonstrating more reliable tool-conditioned evidence localization and verification.
Overall, the toolbox-enabled results show that effective long-video grounding requires learned, verification-driven tool use rather than tool availability alone.

\subsection{Ablations on Fidelity-Weighted Advantage Modulation}

\begin{figure}[!t]
    \centering
    \includegraphics[width=0.9\columnwidth]{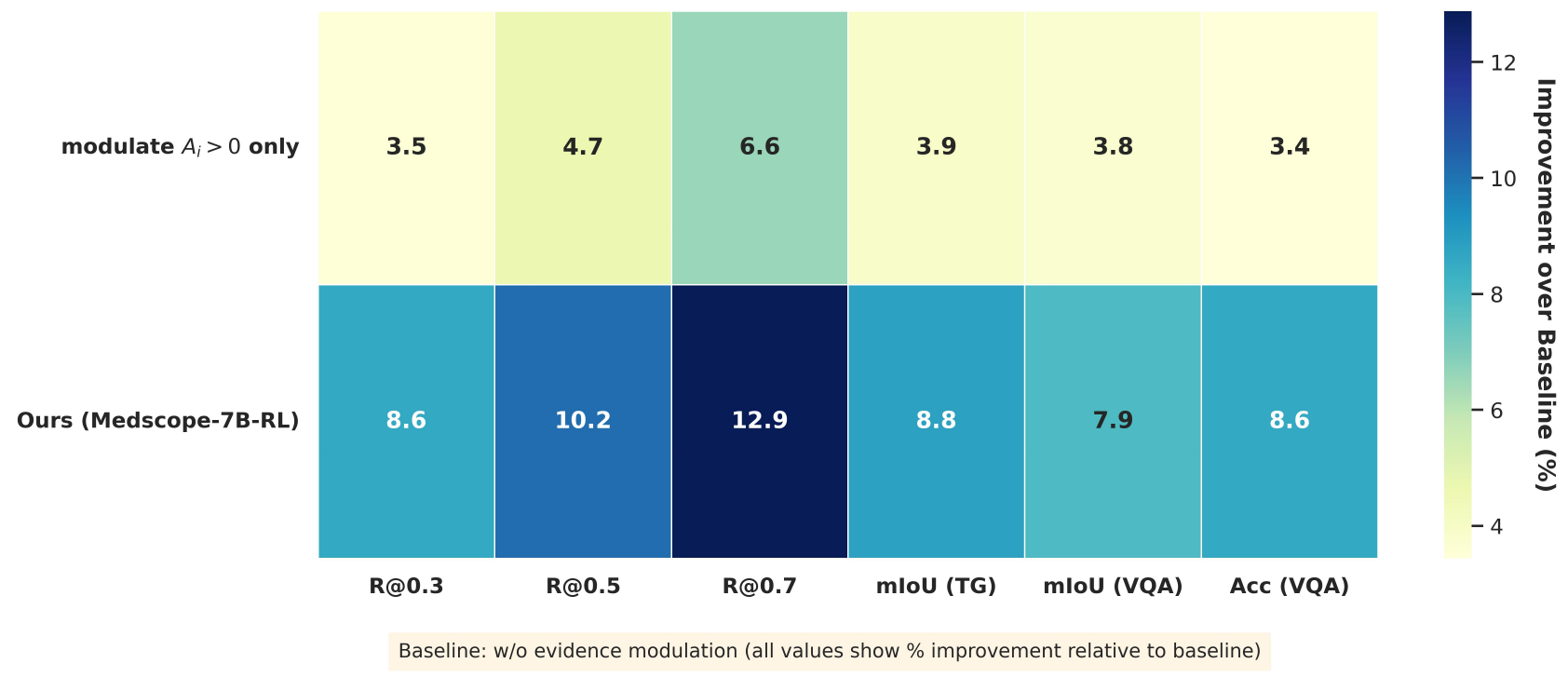}
    \caption{
    Comparison of advantage modulation strategies relative to the baseline.
    \textbf{modulate $A_i>0$ only} applies modulation only to positive-advantage trajectories.
    The baseline is \textbf{w/o evidence modulation}.
    }
    \vspace{-10pt}
    \label{fig:fig6}
\end{figure}

\noindent\textbf{Fidelity-weighted advantage modulation plays a key role in stabilizing grounded policy optimization.}
Figure~\ref{fig:fig6} compares advantage modulation strategies relative to the baseline without evidence modulation.
Applying fidelity-weighted modulation to both positive and negative advantages (\textbf{Ours}) consistently outperforms positive-only modulation.
In particular, full modulation yields larger gains in temporal grounding, improving R@0.7 by 12.9\% compared to 6.6\%, and achieves higher mIoU gains (8.8\% vs.\ 3.9\%).
Similar improvements are observed on grounded VQA, with higher gains in both accuracy (8.6\% vs.\ 3.4\%) and localization quality.

\subsection{Training Dynamics}
\label{app:training_dynamics}

\begin{figure*}[!h]
    \centering
    \includegraphics[width=0.92\linewidth]{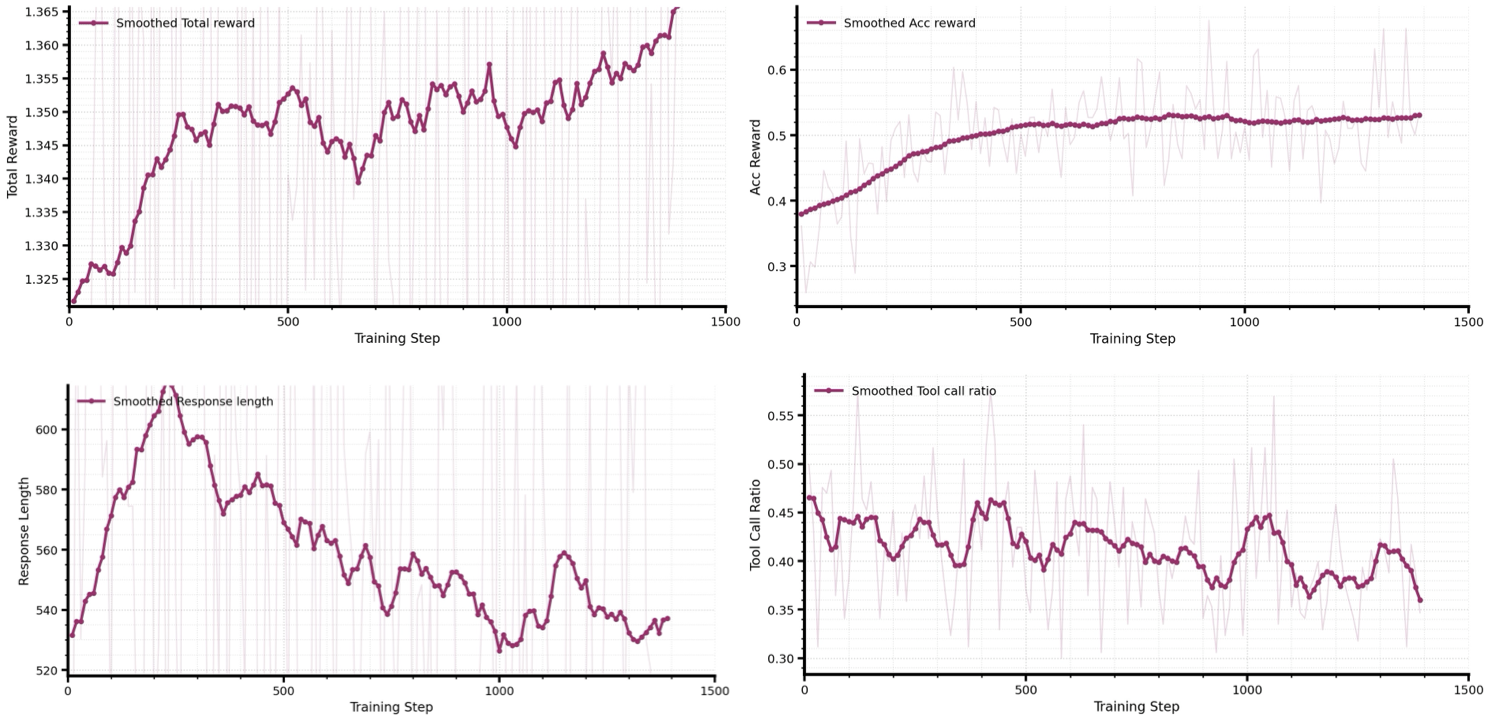}
    \caption{Training dynamics of GA-GRPO. We plot the smoothed total reward, accuracy reward, response length, and tool-call ratio over training steps. Light curves show per-step values, while dark curves show smoothed trends, highlighting stable reward growth, early accuracy saturation, and a gradual shift toward more concise and selective tool use.}
    \label{fig:training_dynamics}
\end{figure*}

Figure~\ref{fig:training_dynamics} visualizes key training signals during GA-GRPO. The total reward shows a steady upward trend with moderate oscillations, indicating stable optimization without late-stage collapse. The accuracy-related reward increases rapidly in the early phase and then gradually plateaus, suggesting that the policy first learns to satisfy answer-level objectives before focusing on finer improvements. Meanwhile, the response length rises at the beginning and then decreases over training, consistent with the model learning more concise reasoning once tool-based evidence becomes reliable. Finally, the tool-call ratio exhibits a mild downward drift with intermittent spikes, implying that the policy becomes more selective in invoking tools while still calling them when evidence is needed. Overall, these dynamics support that GA-GRPO improves both correctness and evidence efficiency, while keeping the rollout behavior stable.

\section{Additional Benchmarks}
\label{appendix:Benchmarks}
\paragraph{SVU-31K}
SVU-31K \citep{wang2025surgvidlm} is a large-scale benchmark designed for multi-grained medical video understanding. It consists of two major task families.
(1) \textbf{Multi-grained Video Description}, which includes Full Video Description and Fine-grained Video Description. These tasks evaluate the model’s ability to generate comprehensive and detailed clinical video summaries across four clinical dimensions, covering content correctness, detail richness, contextual understanding, and temporal coherence.
(2) \textbf{Visual Reasoning}, which comprises Fine-grained Temporal Visual Reasoning and Fine-grained Perception Visual Reasoning. These tasks assess the model’s capability to reason over temporal event sequences and fine-grained visual cues within medical procedures.
\paragraph{ClinVideo-eval}
ClinVideo-eval is a grounded medical video benchmark constructed to evaluate both in-domain and out-of-domain generalization. It includes three datasets: OphVL, MedVideoCap, and SurgVidLM. 
OphVL and MedVideoCap are treated as in-domain benchmarks, while SurgVidLM serves as an out-of-domain test set. 
The former two support both Temporal Grounding and Grounded VQA tasks, whereas SurgVidLM focuses exclusively on Grounded VQA. 
Temporal Grounding evaluates the model’s ability to localize clinically relevant temporal segments, while Grounded VQA assesses joint visual grounding and answer correctness. 
This benchmark enables a comprehensive analysis of model robustness and generalization across diverse medical video domains.

\section{Additional Implementation Details}
\label{appendix:Implementation Details}

\paragraph{Model settings for data synthesis.}
We specify the models and selection hyperparameters used in our three-stage data synthesis pipeline. In Stage 1, the segmenter $\Phi$ is instantiated with \texttt{gemini-2.5-pro} to predict entity-guided windows, clip-level dense captions are produced by \texttt{Qwen3-VL-235B-A22B-Instruct}, and caption merging plus global summarization are performed by \texttt{gpt-4.1}. For clip-level segments, we additionally adopt a lightweight captioning branch that sparsely samples frames within each window and queries \texttt{Qwen2.5-VL-72B} to obtain auxiliary captions. In Stage 2, the cross-model filtering pool $\mathcal{M}$ consists of \texttt{gemini-2.5-flash}, \texttt{grok-4-1-fast}, and \texttt{Qwen3-235B-A22B-Instruct}; we set $\theta_{\text{text}}=1$ and $\theta_{\text{sum}}=1$ to remove candidates answerable without localized evidence, and $\theta_{\text{loc}}=2$ to retain candidates with majority agreement under the local caption view $\mathcal{C}(V;w)$. After text filtering, we perform multimodal confirmation with \texttt{gemini-2.5-flash} by answering from the corresponding video segment $\mathrm{Clip}(V,w)$ and keeping only pairs that match the target answer. In Stage 3, we use \texttt{gemini-2.5-pro} as the teacher to construct environment-interactive VCoT trajectories by iteratively invoking native tools for temporally targeted observations before generating the final prediction.

\begin{table}[htbp]
\centering
\small
\setlength{\tabcolsep}{8pt}
\renewcommand{\arraystretch}{1.15}
\begin{tabular}{lcc}
\toprule
Component & SFT & RL \\
\midrule
Optimizer & AdamW & AdamW \\
Learning Rate (LR) & 5e-5 & 1e-6 \\
LR Scheduler & cosine & constant \\
Weight Decay & 0.0 & 1e-2 \\
No.\ of Training Steps & 1200 & 1327 \\
No.\ of Warmup Steps & 120 & 0 \\
Max Length & 51200 & 52384 \\
Dynamic Batch Size & True & False \\
Remove Padding & True & True \\
Liger Kernel & True & False \\
Per-device train batch size & 2 & - \\
Gradient accumulation steps & 1 & - \\
Rollout engine & - & SGLang \\
Rollouts per prompt ($n$) & - & 16 \\
No.\ of GPUs & 32 & 64 \\
No.\ of Frames & 512 & 512 \\
\bottomrule
\end{tabular}
\caption{Key hyperparameters for supervised fine-tuning and reinforcement learning.}
\label{tab:hparams_key}
\end{table}

\paragraph{SFT.}
We implement SFT with lmms-engine using an FSDP2 trainer (Table~\ref{tab:hparams_key}). We train in BF16 and enable FlashAttention-2 for efficient long-context attention. Input sequences are packed to a maximum length of 51{,}200 tokens with a first-fit strategy, overlong samples are filtered, and dynamic batch sizing is enabled to maximize throughput under variable video and text lengths. To reduce memory usage, we enable gradient checkpointing and remove padding, and we further enable the Liger kernel. Videos are loaded through the qwen\_vl\_utils backend with an FPS-based sampling strategy, capped at 512 frames, and constrained by a maximum pixel budget of 50{,}176. Optimization uses AdamW with a learning rate of $5\times10^{-5}$, weight decay $0.0$, a cosine scheduler, and 120 warmup steps. We use a per-device batch size of 2 with gradient accumulation 1, train for 1{,}200 steps, and run on 32 H200 GPUs. Checkpoints are saved every 200 steps with a maximum of 5 retained. This SFT stage takes approximately 22.5 hours on 32 GPUs.

\paragraph{RL.}
We implement RL with verl and generate rollouts using SGLang (Table~\ref{tab:hparams_key}). We disable entropy regularization by setting the actor entropy coefficient to 0 and optimize with AdamW at a learning rate of $10^{-6}$ under a constant learning-rate schedule with weight decay $10^{-2}$. Training runs for 1{,}327 steps with no warmup and a maximum sequence length of 52{,}384 tokens. We enable gradient checkpointing for the actor model and use BF16 mixed precision throughout, explicitly setting reduce and buffer dtypes to BF16. Both the actor and reference models are trained under FSDP with parameter and optimizer offloading disabled to avoid host-device transfer overhead. For rollouts, we set the rollout engine to sglang with tensor model parallel size 1, log-prob micro-batch size per GPU 1 for both actor and reference, and we cap rollout GPU memory utilization at 0.4 to maintain stable serving during sampling. We sample $n=16$ rollouts per prompt and use Ulysses sequence parallelism with size 4 for the actor. We further remove KL from the reward matching our GA-GRPO training setup. RL is trained on 64 H200 GPUs.

\section{Additional Baselines \& Metrics}
\label{appendix:Baselines & Metrics}

\subsection{Baseline Models}
We compare MedScope with several baseline methods, which are divided into proprietary models, open-source models, reasoning models, and reasoning agents.

\paragraph{Proprietary Models}
This category includes state-of-the-art closed-source models such as GPT-4o \citep{hurst2024gpt}, Gemini 3-Pro-Preview \citep{comanici2025gemini}, and Gemini-2.5-Flash \citep{comanici2025gemini}.

\paragraph{Open-source Models}
The open-source baselines include video-language models such as Qwen2.5-VL-7B \citep{bai2025qwen2}, InternVL3-8B \citep{zhu2025internvl3}, VideoLLaMA3-7B \citep{zhang2025videollama}, Video-LLaVA-7B \citep{lin2024video}, GroundingGPT-7B \citep{lin2024video}, and SurgVidLM \citep{wang2025surgvidlm}.

\paragraph{Reasoning Models}
Reasoning-oriented baselines include VideoR1-7B \citep{feng2025video}, VideoChat-R1-7B \citep{li2025videochat}, and VideoRFT-7B \citep{wang2025videorft}.

\paragraph{Reasoning Agents}
Reasoning agents incorporate structured tool use and multi-step inference, including MedScope-7B-SFT, MedScope-7B-RL, ReWatch-R1-7B \citep{zhang2025rewatch}, and LongVT-7B-RFT \citep{yang2025longvt}.

\subsection{Evaluation Metrics}
For video description tasks, we follow the evaluation pipeline in SurgVidLM \citep{wang2025surgvidlm}, employing a GPT-based evaluator to score model outputs on a 1--5 scale across four clinical aspects: correctness of information (CI), detail orientation (DO), contextual understanding (CU), and temporal understanding (TU).
For SVU-31K, we additionally report standard captioning and reasoning metrics, including BLEU-4 \citep{papineni2002bleu}, CIDEr \citep{vedantam2015cider}, METEOR \citep{banerjee2005meteor}, and ROUGE-L \citep{lin2004rouge}.
For grounded medical video understanding, Temporal Grounding is evaluated using Recall at multiple IoU thresholds and mIoU, while Grounded VQA is evaluated using mIoU and accuracy.

\section{Case Analysis}
\label{appendix:case}

This section provides brief case studies to visualize test-time tool use. We highlight recurring behaviors learned after VCoT cold-start and GA-GRPO, showing how the model localizes evidence with \texttt{crop\_video}, probes details with \texttt{get\_frame}, and composes them for verification. We also include representative failure cases, revealing residual phenomena where negative evidence does not always trigger re-localization and where textual priors can occasionally dominate fine-grained visual binding.

\subsection{Good Cases Analysis}

\paragraph{Hypothesis-Driven Temporal Localization.}
Figure~\ref{fig:good_case_1} illustrates an emergent test-time behavior after VCoT cold-start and GA-GRPO: the model separates hypothesis formation from evidence acquisition, and uses temporally targeted tool calls as a verification step rather than an optional add-on. Instead of answering immediately from coarse context, it first proposes a plausible surgical rationale with explicit temporal anchoring, then commits to a \texttt{crop\_video} call that concentrates computation on the suspected decision window. The second turn shows a stable evidence integration loop where the retrieved clip is treated as the decisive signal to confirm or revise the hypothesis before producing the final answer. This suggests the model has learned a lightweight “plan then act then verify” policy with calibrated stopping, which is difficult to elicit from tool access alone and aligns with the intended training objective of coupling reasoning with temporally grounded verification.

\paragraph{Framewise Detail Probing.}
Figure~\ref{fig:good_case_2} highlights a characteristic behavior learned after tool-augmented training: the model treats \texttt{get\_frame} as a targeted visual probe for fine-grained attributes, using minimal queries to resolve a localized detail question instead of over-retrieving. A notable trait is its failure-aware control policy: when a probe violates the clip boundary and the tool returns an execution error, the model does not collapse into speculation, but reuses already validated temporal evidence and preserves the queried action context to complete the answer. This indicates improved robustness to tool noise and boundary conditions, as well as an internalized notion of evidence sufficiency for short-clip decision making.

\paragraph{Fine-to-Coarse Evidence Search.}
Figure~\ref{fig:good_case_3} illustrates an emergent control behavior after tool-use training: the model treats temporal retrieval as an iterative search problem and adjusts the evidence window based on observed insufficiency. Instead of committing to the first hypothesis-driven crop, it diagnoses that the initial narrow interval lacks decisive evidence, then deliberately broadens the crop to capture the complete procedural sequence and any on-screen cues. This reflects a learned notion of evidence completeness and a self-correcting retrieval policy, improving reliability when the first tool call under-covers the critical moment.

\paragraph{Parallel Time-Jump Self-Correction.}
Figure~\ref{fig:good_case_4} reflects an emergent failure-aware retrieval strategy learned through tool-use training. The first crop retrieves a segment that is visually well formed but semantically misaligned with the question because it remains in the speaker presentation rather than the operative field. The model explicitly recognizes this mismatch and treats it as an evidence acquisition error instead of forcing an answer from irrelevant frames. It then revises its temporal hypothesis and performs a parallel jump to a distant time window that is more likely to contain the suturing step. After retrieving the corrected interval, the model grounds the answer in a concrete spatial cue by locating the needle entry at the junction where the two wound edges meet. This behavior indicates that the model has learned to decouple retrieval from decision making, using tool feedback to correct localization and improve reliability when early evidence windows are wrong.

\paragraph{Coarse-to-Fine Tool Chaining.}
Figure~\ref{fig:good_case_5} demonstrates an emergent coarse-to-fine verification behavior where the model composes multiple tools into a single evidence-seeking routine rather than treating each call independently. The model begins with a deliberately wide crop to maximize recall because the target event is temporally sparse and easy to miss under motion, haze, and intermittent occlusion. After detecting a plausible two-tool overlap, it reduces the temporal extent to suppress distractors and stabilize the visual context, effectively trading coverage for precision once a candidate window has been identified. It then switches tools from temporal localization to frame-level inspection, selecting a representative timestamp inside the overlap to make the count checkable from a single image and therefore less sensitive to narrative drift. This chaining behavior suggests the model has learned an internal division of labor across tools, using cropping for search and get\_frame for confirmation, which improves robustness for brief concurrency queries that would otherwise be unreliable from global summarization alone.

\subsection{Failure Cases Analysis}

\paragraph{Evidence-Overrule Failure.}
Figure~\ref{fig:bad_case_1} shows that the model can invoke a tool for verification and explicitly report that the retrieved frames do not contain the queried action. The trajectory then transitions to an answer without launching a follow-up localization step, suggesting a weak linkage between negative evidence and re-planning. One plausible contributing factor is that the policy has learned a common response template where a single verification attempt is often sufficient, while cases requiring iterative recovery from uninformative crops are less emphasized. This leads to occasional reliance on procedural priors when tool feedback is inconclusive.

\paragraph{Text-Anchored Entity Binding Drift.}
Figure~\ref{fig:bad_case_2} shows that the model can use a tool to retrieve a broad segment that mixes case text with the operative view, and it then commits to an anatomical label that is strongly conditioned on the textual anchor. The trajectory suggests that the case description is treated as a high-confidence prior, while the fine-grained visual binding between the instrument and the grasped tissue is not explicitly stress-tested. This can yield occasional confusion between the true target structure and a nearby landmark when the target is small or visually ambiguous.

\section{Prompts}
\label{appendix:prompts}

\subsection{Training Prompts}
\begin{figure}[t]
    \centering
    \includegraphics[width=0.9\linewidth]{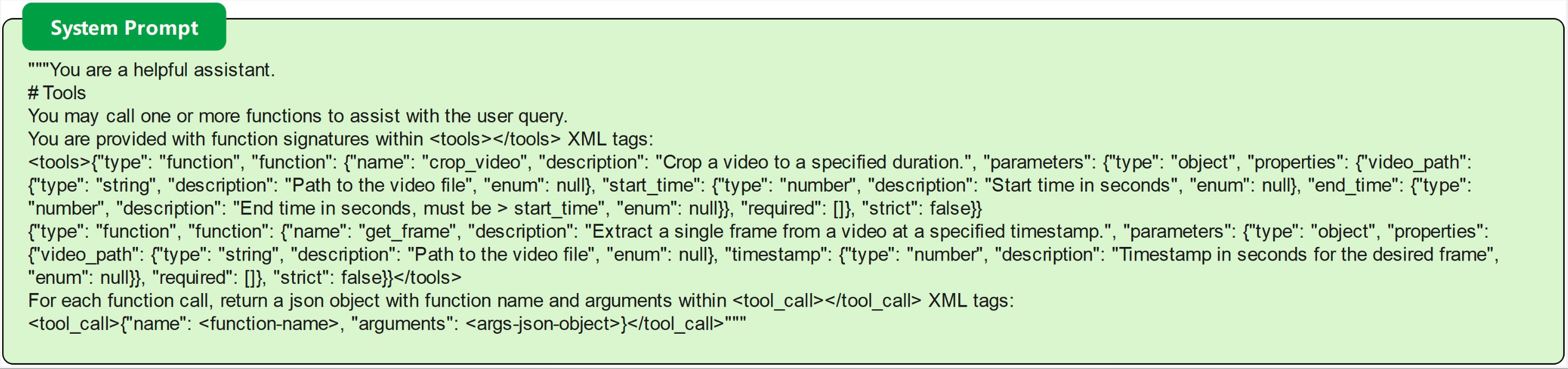}
    \caption{System prompt of MedScope.}
    \label{fig:sys_prompt}
\end{figure}

\begin{figure}[t]
    \centering
    \includegraphics[width=0.9\linewidth]{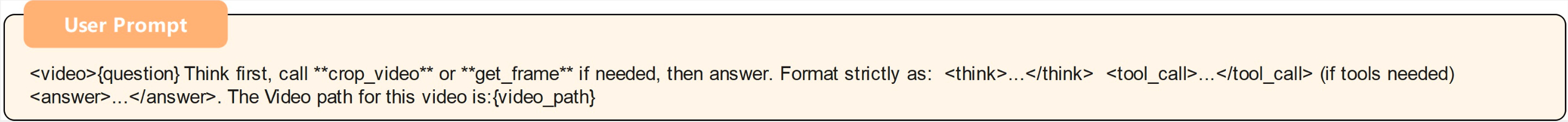}
    \caption{User prompt of MedScope.}
    \label{fig:usr_prompt}
\end{figure}

To operationalize the coarse-to-fine clinical reasoning framework (Section~\ref{sec:coarse_to_fine_reasoning_framework}), MedScope employs a lightweight yet \emph{strictly structured} prompting interface that couples hypothesis-driven textual reasoning with temporally targeted evidence acquisition.
As shown in Figure~\ref{fig:sys_prompt} and Figure~\ref{fig:usr_prompt}, our prompts are designed to (i) expose a compact visual toolbox through an explicit function contract, (ii) enforce an unambiguous interaction protocol for multi-round trajectories, and (iii) standardize output boundaries to make tool use and final predictions reliably parsable and verifiable.

\paragraph{System prompt (tool contract and executable interface).}
Figure~\ref{fig:sys_prompt} presents the system prompt, which defines the assistant role and specifies the available tools via machine-readable function signatures enclosed in \textless tools\textgreater...\textless/tools\textgreater.
This contract-first design makes tool invocation an explicit, executable action rather than an implicit textual suggestion, thereby supporting inspectable trajectories of the form $\tau_i=\{(t_{i,k},a_{i,k},o_{i,k})\}_{k=1}^{K_i}$ (Section~\ref{sec:coarse_to_fine_reasoning_framework}).
Concretely, the action space contains two native functions that implement hierarchical temporal inspection:
(1) \texttt{crop\_video}, which takes a video path and a coarse time window and returns a clip for localized, denser evidence gathering;
and (2) \texttt{get\_frame}, which takes a video path and a fine-grained timestamp to extract frames around a specific moment for verification.
This coarse-to-fine evidence retrieval pattern aligns with prior tool-augmented video reasoning paradigms that dynamically sample additional frames on demand to reduce hallucination and improve long-video grounding.

\paragraph{User prompt (goal specification and protocol constraints).}
Figure~\ref{fig:usr_prompt} shows the user prompt template, which injects the instance query and the concrete video path while enforcing a strict output protocol:
the model must first produce textual rationalization in \texttt{\textless think\textgreater...\textless/think\textgreater}, optionally emit one or more tool calls in \texttt{\textless tool\_call\textgreater...\textless/tool\_call\textgreater} (when evidence is needed), and finally output the terminal prediction in \texttt{\textless answer\textgreater...\textless/answer\textgreater}.
By explicitly separating \emph{reasoning} (\texttt{<think>}) from \emph{evidence acquisition} (\texttt{<tool\_call>}) and \emph{final decision} (\texttt{<answer>}), the prompt makes the decision process externally auditable and naturally supports iterative verification, where observations returned by the environment are appended back into context for the next round (Section~\ref{sec:coarse_to_fine_reasoning_framework}).
Moreover, the prompt encourages \emph{evidence-seeking} behavior (invoke \texttt{crop\_video}/\texttt{get\_frame} ``if needed''), which is critical for clinical scenarios where temporally localized cues must be checked rather than assumed.

\begin{figure}[t]
    \centering
    \includegraphics[width=0.9\linewidth]{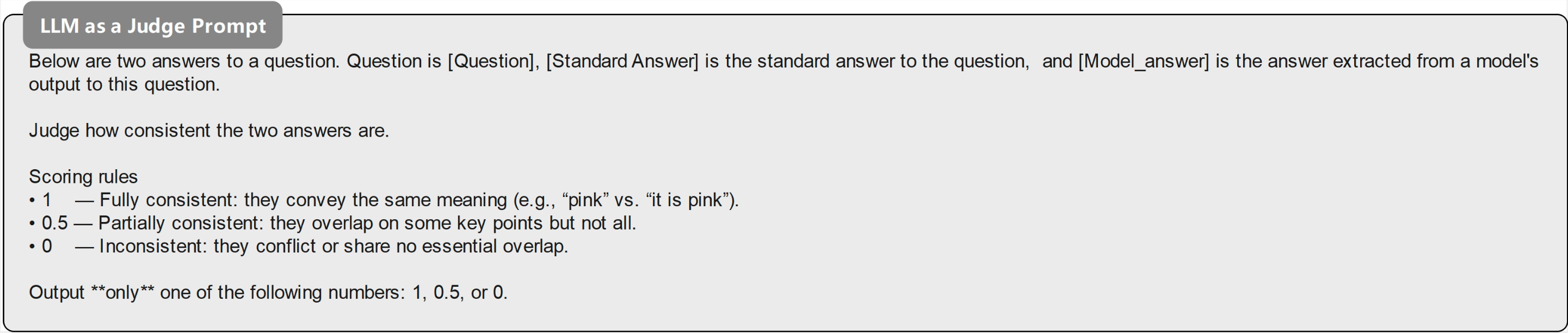}
    \caption{Prompt for LLM-as-a-Judge.}
    \label{fig:judge_prompt}
\end{figure}

\paragraph{LLM-as-a-Judge prompt.}
As shown in Figure~\ref{fig:judge_prompt}, we adopt a lightweight LLM-as-a-judge template to measure the semantic consistency between two answers: the reference and the extracted model output given the same Question.
The judge is instructed to output only one scalar from $\{1,\,0.5,\,0\}$, corresponding to fully consistent, partially consistent, and inconsistent.
This discrete, format-constrained scoring avoids verbose rationales and reduces parsing ambiguity, enabling robust aggregation (e.g., voting or consensus) when computing agreement signals in our filtering pipeline.

\subsection{Prompting Details for ClinVideoSuite Data Synthesis}

\textbf{ClinVideoSuite} is constructed to be evidence-centric and temporally localized: captions, QA pairs, and VCoT trajectories are all tied to explicit supervision windows and are constrained to rely on visually observable cues.
To make this scalable and automatically verifiable, we use a set of strict, interface-oriented prompts that standardize (i) temporal segmentation, (ii) dense clip captioning and global summarization, (iii) window-grounded QA construction, and (iv) environment-interactive VCoT trajectory synthesis.

\subsubsection{Stage 1: Evidence-centric captioning and global summarization.}
Stage~1 builds timestamped dense captions as localized evidence and a compact global summary as context.
This stage is implemented by three prompts that form a coarse-to-fine summarization stack: semantic boundary detection $\rightarrow$ clip-level dense captioning $\rightarrow$ chronological caption merging.

\paragraph{Original Clip Prompt: semantic boundary detection for windowing.}
As shown in Figure~\ref{fig:prompt_clip_origin}, given sparse, chronological frames from a video interval (each frame annotated with an absolute timestamp in seconds), the prompt asks the model to propose semantic change points that split the interval into coherent phases or steps.
Crucially, it enforces a JSON-only output with a deterministic schema:
\texttt{\{"cut\_points":[...], "segment\_summaries":[...]\}},
where the summaries are \emph{single concise phrases} and the segments must be continuous and cover the full interval.
The prompt further imposes structural constraints that stabilize window generation at scale: cut points must be strictly increasing and lie inside the open interval (excluding boundaries), a minimum number of segments is required for sufficiently long clips (unless truly no semantic change exists), and very short segments are discouraged via an explicit minimum-length constraint.
These rules make $\mathcal{B}(V)$ (Eq.~(2)) \emph{well-formed} and prevent degenerate partitioning (e.g., missing coverage or noisy micro-segments), yielding reliable supervision windows for downstream dense captioning and QA grounding.

\paragraph{Clip Caption Prompt: dense, clinically grounded, \emph{visual-only} descriptions.}
For each predicted window $w=[s,e]$, the model generates a dense caption that describes only what is visually supported within that temporal span.
The prompt in Figure~\ref{fig:prompt_clip_caption} explicitly prioritizes medical and surgical semantics---anatomy or target tissue, instruments, actions, and workflow steps---and encourages spatial relations when visible (e.g., left/right, proximal/distal, superficial/deep, anterior/posterior).
At the same time, it includes a strong anti-hallucination constraint that forbids inventing patient information, diagnoses, or outcomes not directly observable.
This balances two needs: (i) high-fidelity evidence that captures fine-grained visual cues for grounding, and (ii) conservative phrasing that reduces caption-induced shortcuts in later QA generation and verification.

\paragraph{Caption Merge Prompt: compressing chronological caption blocks into a global summary.}
To obtain a compact global context $S(V)$ without leaking localized cues, we apply the prompt in Figure~\ref{fig:prompt_clip_merge} to merge two chronological caption blocks into one shorter block while preserving timestamp ranges at the beginning of each output line.
The prompt enforces plain-text lines only (no headings, bullets, or markdown), strict chronological order, and aggressive compression that keeps only the surgical/clinical storyline (major steps, key anatomy, and essential instruments).
It explicitly drops generic filler (e.g., emotions, professionalism, operating-room environment narration, long instrument inventories), merges adjacent lines describing the same step, caps the output length (e.g., $\leq 6$ lines), and forbids introducing new facts.
As a result, $S(V)$ remains a coarse overview that supports high-level reasoning, while preserving the necessity of retrieving temporally localized evidence from $\mathcal{C}(V)$ for fine-grained grounding.

\subsubsection{Stage 2: High-quality QA construction with localized evidence dependence.}
Stage 2 constructs QA pairs that hinge on temporally localized visual cues and are not answerable from coarse context alone.
As shown in Figure~\ref{fig:qa_prompt}, we operationalize this objective with a single, strongly constrained prompt that explicitly separates global context from local evidence.

\paragraph{QA Pair Generation Prompt: clip-answerable but global-not-answerable QA.}
The QA pair generation prompt takes as input:
(A) a \emph{Global Summary} (coarse overview of the full video),
(B) a \emph{Clip Caption} (local window description with fine-grained details),
(C) the \emph{Clip Timecodes} $[s,e]$ in seconds, and
(D) an upper bound $K$ on the number of QA pairs.
It then generates between $1$ and $K$ QA items under eight hard rules:
\textbf{(i)} one single-part English question (length-limited; avoid double-barreled forms),
\textbf{(ii)} \emph{visual-only} evidence (no domain knowledge, guessing, or narration-only facts),
\textbf{(iii)} answerable using the clip caption but not confidently answerable using only the global summary (prefer details present in the clip caption but absent/vague globally),
\textbf{(iv)} concrete and fine-grained (objects/instruments, colors, actions, temporal order, spatial relations; avoid subjective/evaluative questions),
\textbf{(v)} short, definite answers (length-limited; disallow hedging),
\textbf{(vi)} no answer leakage in the question,
\textbf{(vii)} no explicit mentions of ``clip/segment'' or timestamps in the question, while forcing each output item to carry \texttt{start\_time} and \texttt{end\_time} exactly equal to the provided timecodes,
and \textbf{(viii)} diversity across multiple QA items (no paraphrases; cover different visual details).
The output is JSON-only as a list of objects, making it easy to validate automatically and to align each QA with a supervision window for later multimodal confirmation and trajectory synthesis.

\subsubsection{Stage 3: Environment-interactive visual-CoT synthesis.}
Stage~3 synthesizes visual-CoT trajectories by having a teacher model alternate between hypothesis-driven reasoning and tool-based evidence retrieval in a real video environment.
We implement this with a two-phase VCoT prompting protocol in Figure~\ref{fig:vcot_prompt} that explicitly encodes coarse-to-fine verification and yields inspectable tuples $(t_{i,k}, a_{i,k}, o_{i,k})$.

\paragraph{VCoT Generation Prompts: Phase~1 global skim and planning (Round~1).}
The \textbf{Round~1 VCoT prompt} casts the model as a long-video reasoning assistant and exposes a native tool interface (e.g., \texttt{crop\_video}) via \texttt{\textless tools\textgreater...\textless/tools\textgreater} signatures.
To make tool use executable and parsable, it requires each tool invocation to be wrapped in \texttt{\textless tool\_call\textgreater...\textless/tool\_call\textgreater} and prohibits emitting \texttt{\textless tool\_response\textgreater} directly (the environment injects observations after execution).
The prompt further enforces a strict decision structure: in each round, the model must first write a non-empty, evidence-integrating \texttt{Thinking} section (3--6 sentences, with natural time anchors), then output either exactly one tool call and stop, or a final answer and stop (never mixing both).
This ``plan-then-act'' constraint encourages the teacher to articulate what evidence is missing before querying the environment, aligning with our coarse-to-fine verification objective.

\paragraph{ VCoT Generation Prompts: Phase~2 fine-grained inspection (Round~2 to Round~n).}
For subsequent rounds, the \textbf{fine inspection template} instructs the model to continue the trajectory without repeating earlier content, produce a short reflective \texttt{Thinking} section grounded in the newly retrieved frames (typically low-resolution segment frames), and again choose exactly one action: either request another tool call (one per round) or output the terminal answer.
The prompt explicitly reminds the model not to expose privileged metadata (e.g., time-window hints or reference answers) in its reasoning text, which prevents leakage while still supporting consistent supervision during synthesis.
Overall, the two-phase prompting design induces a natural hierarchy: global skim to localize candidate evidence regions, followed by iterative fine inspection to confirm or revise hypotheses with temporally targeted observations, producing high-fidelity environment-interactive VCoT trajectories suitable for cold-start and RL stages.

\section{Limitations}

While MedScope demonstrates strong evidence-grounded reasoning on long-form clinical videos, several limitations remain. First, our study focuses on 2D endoscopic and surgical video streams with time-localized evidence and does not yet extend to volumetric or spatially registered modalities such as 3D endoscopy, CT, MRI, ultrasound, or multi-view operating-room recordings. As a result, MedScope may not fully capture clinical workflows that require reasoning over three-dimensional anatomy, cross-modal alignment, or longitudinal fusion of imaging with perioperative records. In addition, although our evaluation suite includes physician verification for evidence traceability, the overall scope of expert review is necessarily bounded, and broader validation across more institutions, procedures, and annotation protocols would further strengthen conclusions about generalizability.

Second, the current tool interface is intentionally lightweight, centered on temporal densification and frame-level verification. This design improves reliability and efficiency, but it does not exhaust the range of tools that could benefit clinical decision-making, such as instrument and anatomy segmentation, action phase recognition, OCR for on-screen metadata, structured report generation, or integration with external knowledge bases and patient context. Extending the toolbox to richer modalities and more diverse perception primitives, as well as scaling training and rollouts under larger compute budgets, may further improve robustness, reduce failure cases under ambiguous visual evidence, and enable more comprehensive clinical reasoning beyond video-only settings.

\section{Future Work}

A natural next step is to broaden MedScope from video-only reasoning to holistic multimodal clinical intelligence by incorporating volumetric and multi-view signals such as 3D endoscopy, CT, MRI, and ultrasound, and by aligning video evidence with structured patient context across the perioperative timeline. On the modeling side, extending the action space beyond temporal cropping and frame querying to include richer perception and analysis tools, such as segmentation, tracking, phase recognition, and structured summarization, may enable more precise evidence attribution and stronger robustness under challenging or rare visual patterns. Finally, scaling physician-verified evaluation to a wider range of procedures and institutions, and exploring more principled reward shaping and uncertainty-aware stopping criteria for agentic rollouts, are promising directions to further improve reliability, safety, and generalization for real-world deployment.

\begin{figure*}[h]
    \centering
    \includegraphics[width=\linewidth]{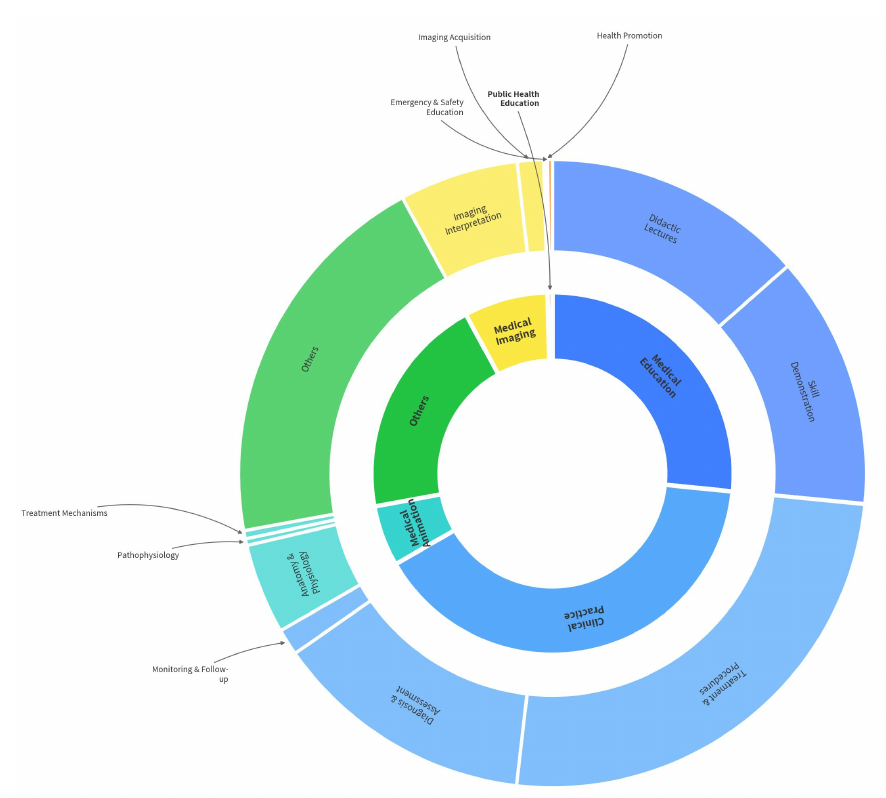}
    \caption{Hierarchical Category Distribution of ClinVideo-QA-254K (MedVideoCap): Inner Ring for Top-Level and Outer Ring for Second-Level Categories.}
    \label{fig:data3}
\end{figure*}

\begin{figure*}[h]
    \centering
    \includegraphics[width=\linewidth]{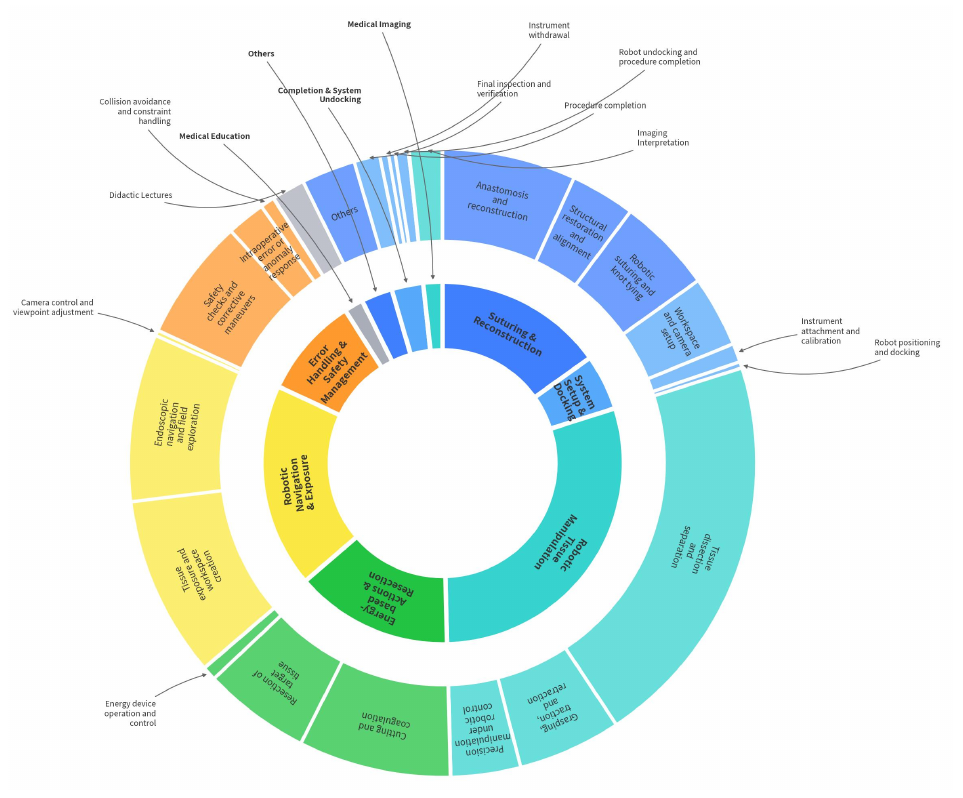}
    \caption{Hierarchical Category Distribution of ClinVideo-QA-254K (SurgVidLM): Inner Ring for Top-Level and Outer Ring for Second-Level Categories.}
    \label{fig:data3-2}
\end{figure*}

\begin{figure*}[h]
    \centering
    \includegraphics[width=\linewidth]{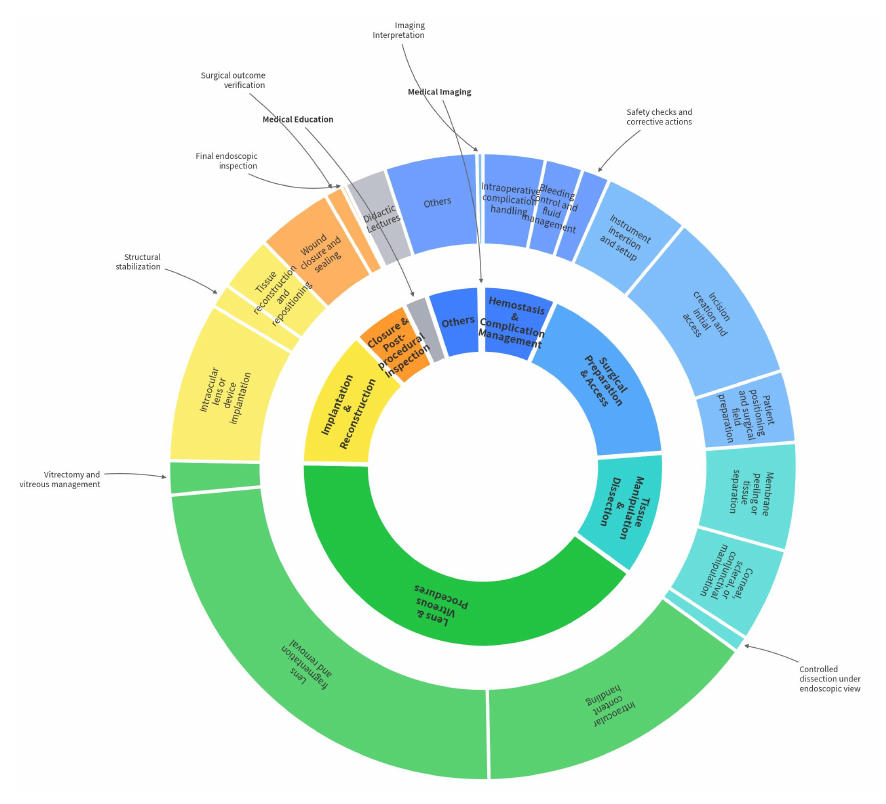}
    \caption{Hierarchical Category Distribution of ClinVideo-QA-254K (OphVL): Inner Ring for Top-Level and Outer Ring for Second-Level Categories.}
    \label{fig:data3-3}
\end{figure*}

\begin{figure}[h]
    \centering
    \includegraphics[width=\linewidth]{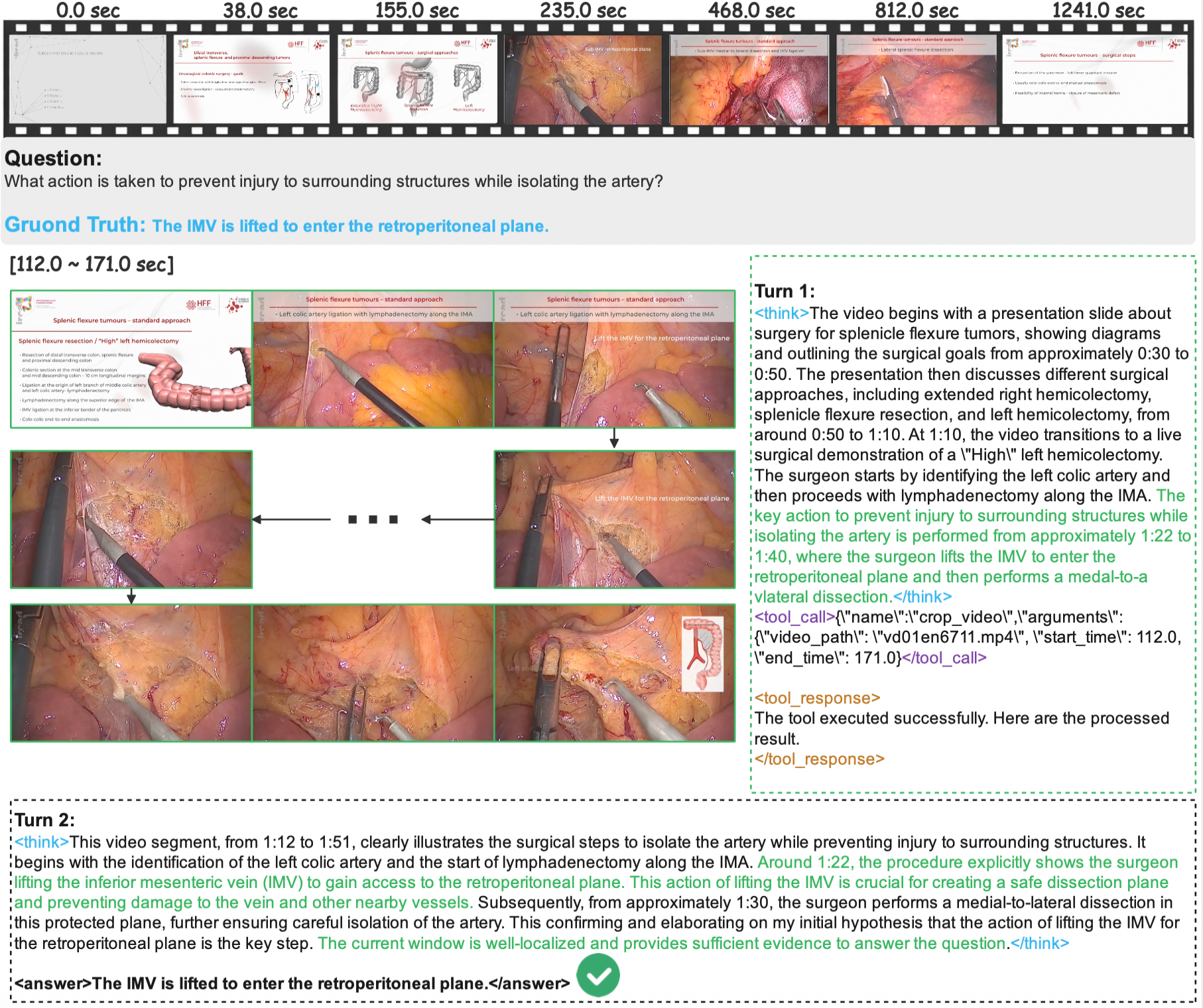}
    \caption{\textbf{Hypothesis-Driven Temporal Localization Example.}
Given an instrument-safety question during arterial isolation, the model first performs a global skim to hypothesize the critical maneuver, then calls \texttt{crop\_video} to zoom into the suspected window (around 1:22--1:40). The retrieved segment verifies that the surgeon lifts the IMV to enter the retroperitoneal plane, enabling safe medial-to-lateral dissection and reducing the risk of injury to surrounding structures. Green highlights indicate key evidence-seeking and verification statements in the VCoT trajectory.}

    \label{fig:good_case_1}
\end{figure}

\begin{figure}[h]
    \centering
    \includegraphics[width=0.9\linewidth]{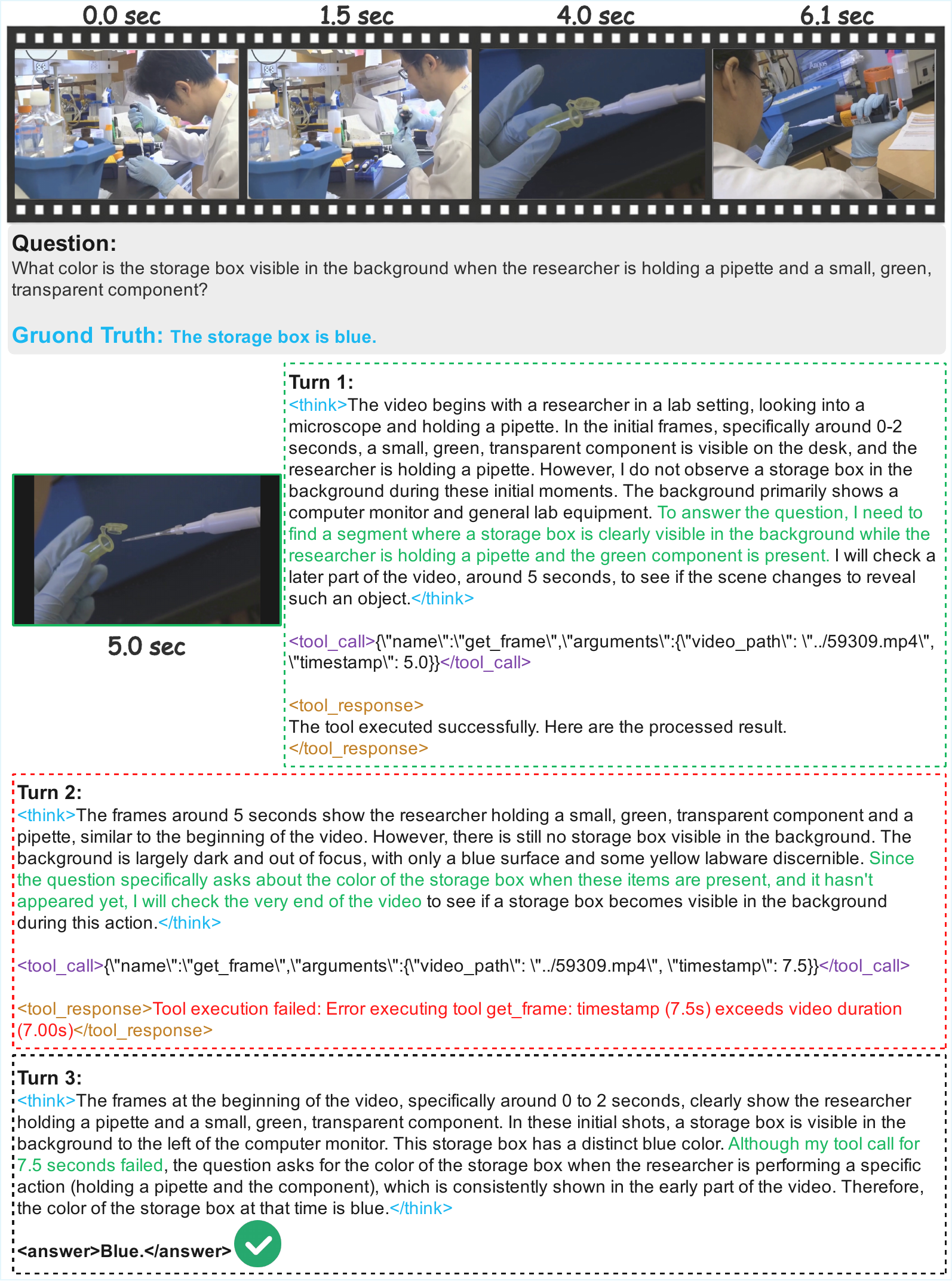}
    \caption{\textbf{Framewise Detail Probing Example.} The model uses \texttt{get\_frame} to probe fine-grained visual attributes in a short clip, first querying a plausible timestamp and then adaptively shifting its probe. When a boundary-violating call fails (timestamp exceeds video duration), it remains robust by falling back to earlier verified frames that satisfy the queried action context, and correctly identifies the storage box color as blue.}

    \label{fig:good_case_2}
\end{figure}

\begin{figure}[h]
    \centering
    \includegraphics[width=0.85\linewidth]{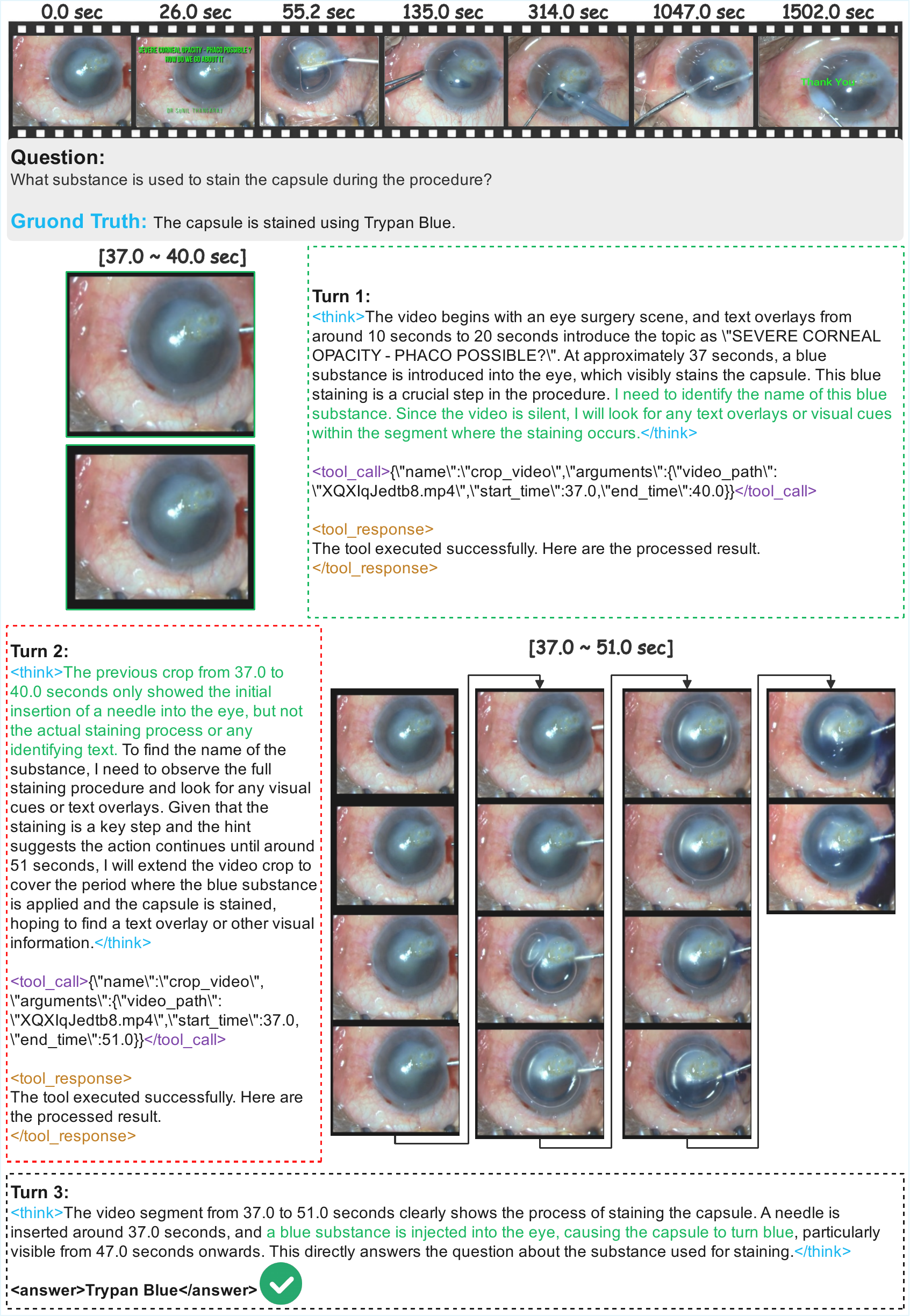}
    \caption{\textbf{Fine-to-Coarse Evidence Search Example.} The model first performs a tight \texttt{crop\_video} around the suspected staining moment, but the narrow window fails to include the decisive staining and any identifying cues. It then adaptively expands to a longer interval that covers the full action, enabling evidence-complete verification and correctly answering that the capsule is stained with Trypan Blue.}

    \label{fig:good_case_3}
\end{figure}

\begin{figure}[h]
    \centering
    \includegraphics[width=\linewidth]{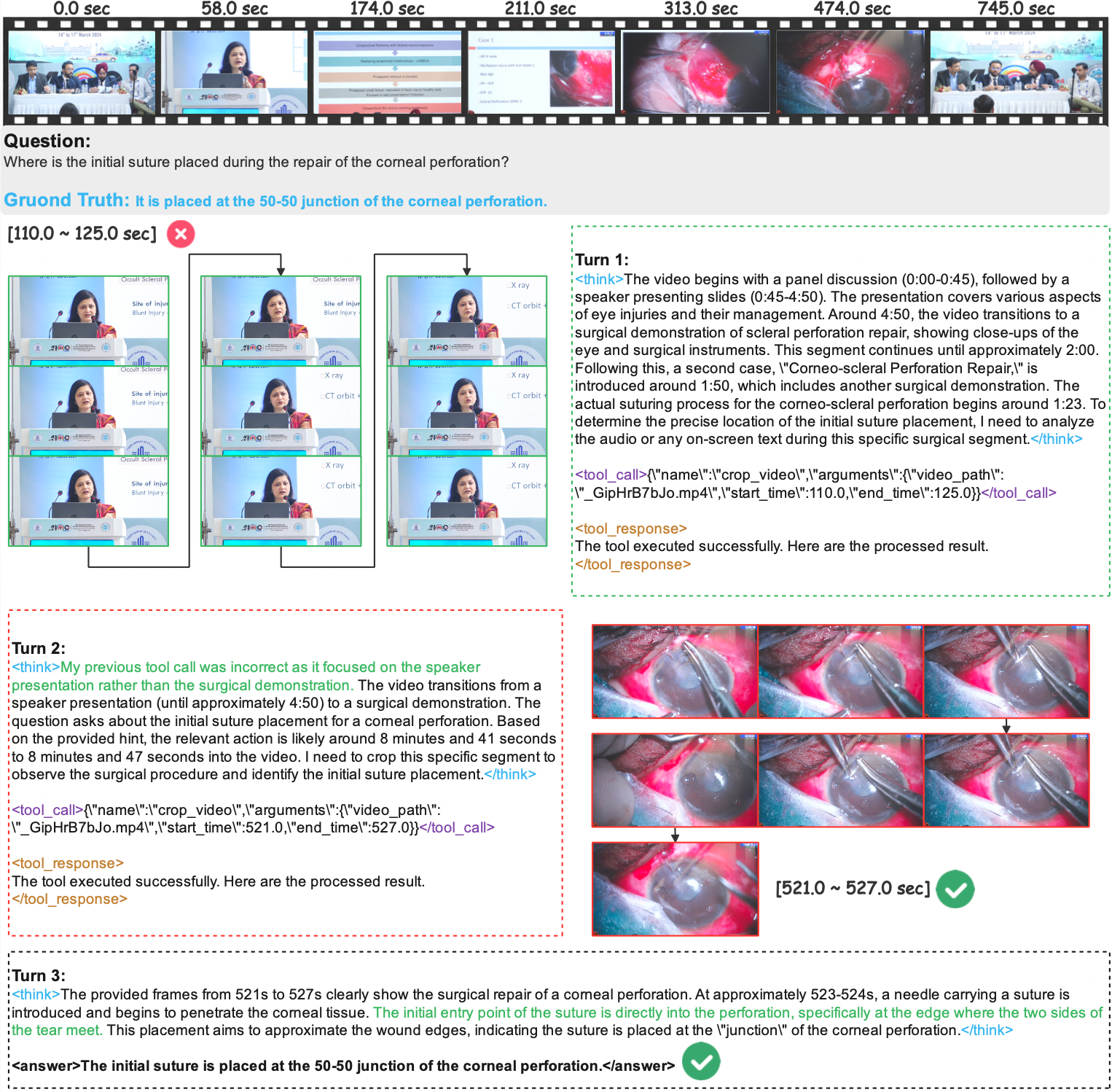}
    \caption{\textbf{Parallel Time-Jump Self-Correction Example.} The model initially crops an irrelevant segment (speaker presentation) and then explicitly recognizes the mismatch between the retrieved evidence and the queried surgical action. It subsequently performs a parallel jump to the correct surgical window and verifies that the first suture is placed at the 50-50 junction where the wound edges meet.}
\label{fig:good_case_4}

\end{figure}

\begin{figure}[h]
    \centering
    \includegraphics[width=\linewidth]{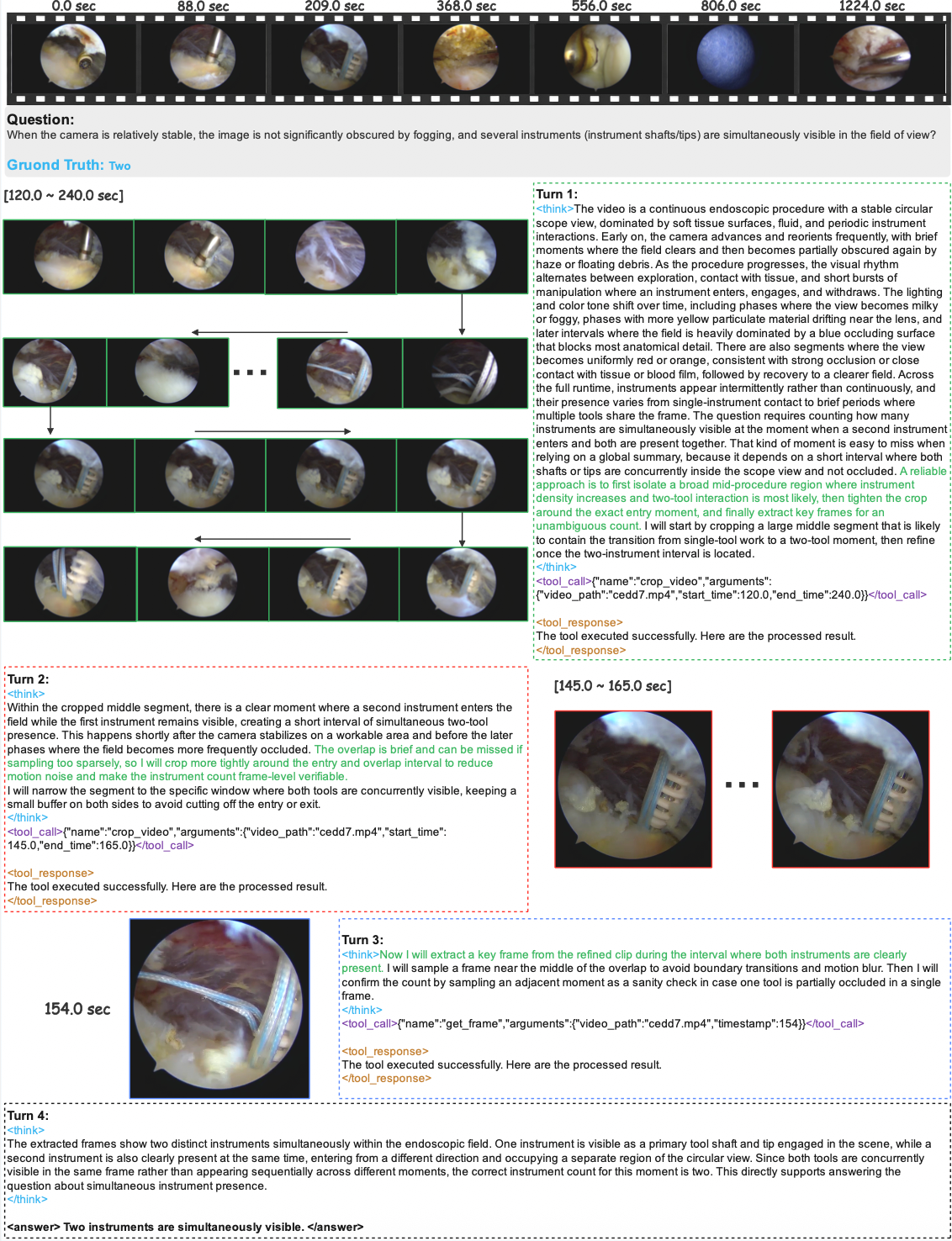}
    \caption{\textbf{Coarse-to-Fine Tool Chaining Example.} The model first performs a broad temporal crop to locate a likely two-instrument overlap, then tightens to a short window around the overlap, and finally uses frame extraction to verify the simultaneous instrument count, yielding the correct answer (two) under stable, low-fog visibility.}
    \label{fig:good_case_5}
\end{figure}

\begin{figure}[h]
    \centering
    \includegraphics[width=\linewidth]{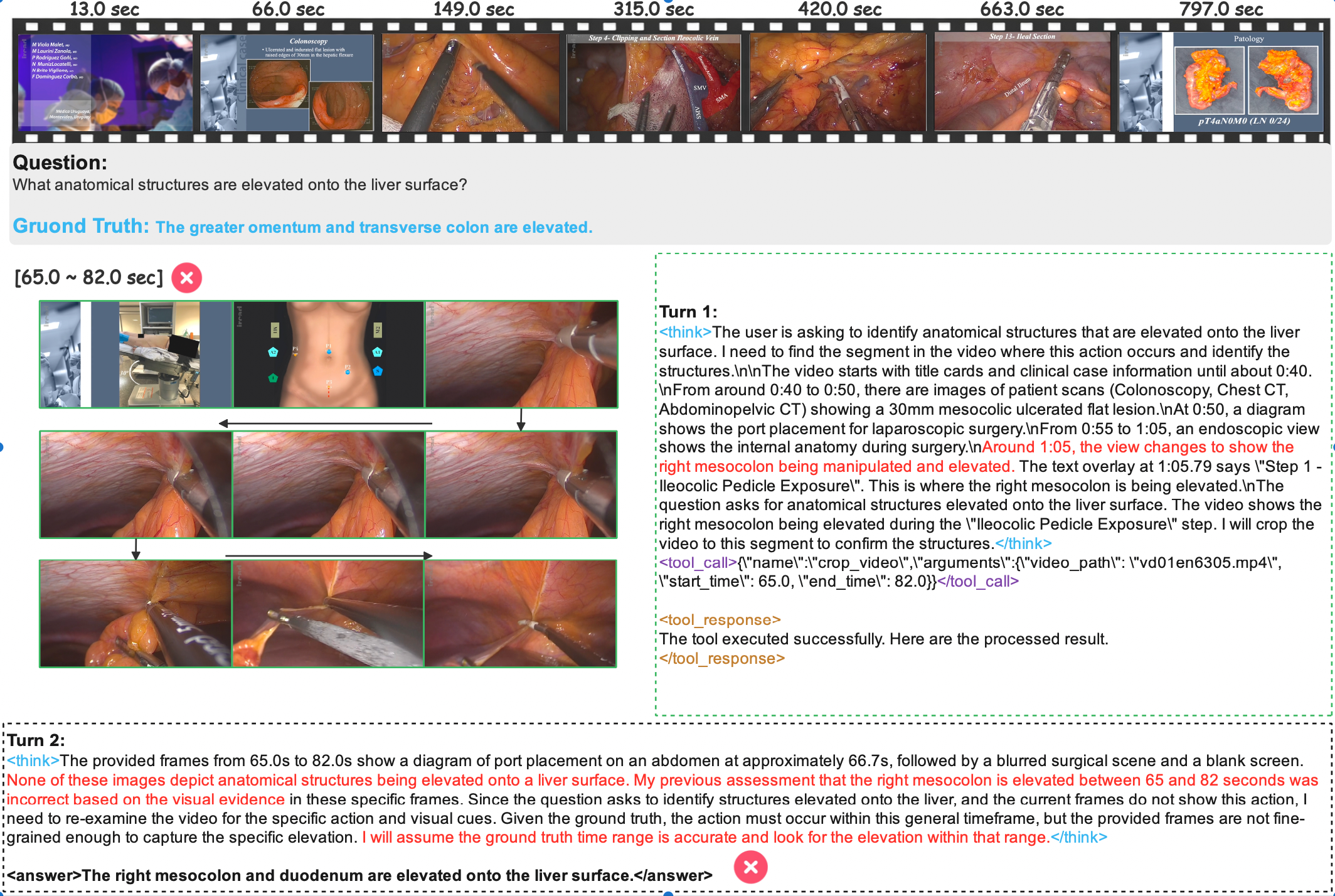}
    \caption{\textbf{Evidence-Overrule Failure example.} The model issues a \texttt{crop\_video} call to verify which structures are elevated onto the liver surface. The returned frames do not show the queried action, and the model notes the mismatch, but it does not proceed with an additional re-localization step and answers based on the prior narrative, yielding an unsupported prediction.}

    \label{fig:bad_case_1}
\end{figure}

\begin{figure}[h]
    \centering
    \includegraphics[width=\linewidth]{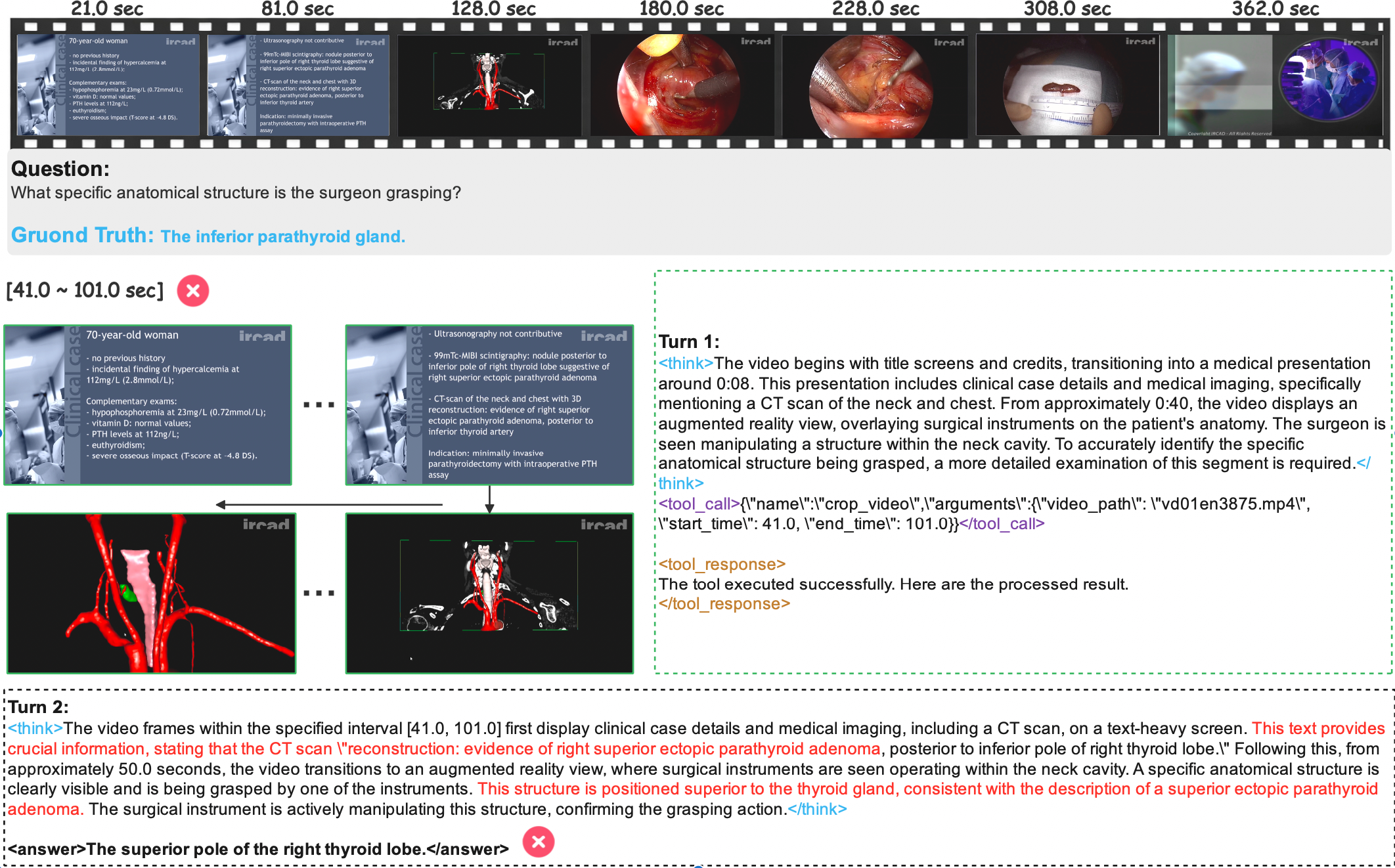}
    \caption{\textbf{Text-Anchored Entity Binding Drift example.} The model crops a long window containing both case text and the operative augmented view, then anchors on the slide description of an ectopic parathyroid lesion and transfers this prior to the grasped target, leading to an incorrect structure identification despite tool use.}

    \label{fig:bad_case_2}
\end{figure}


\begin{figure}[h]
    \centering
    \includegraphics[width=\linewidth]{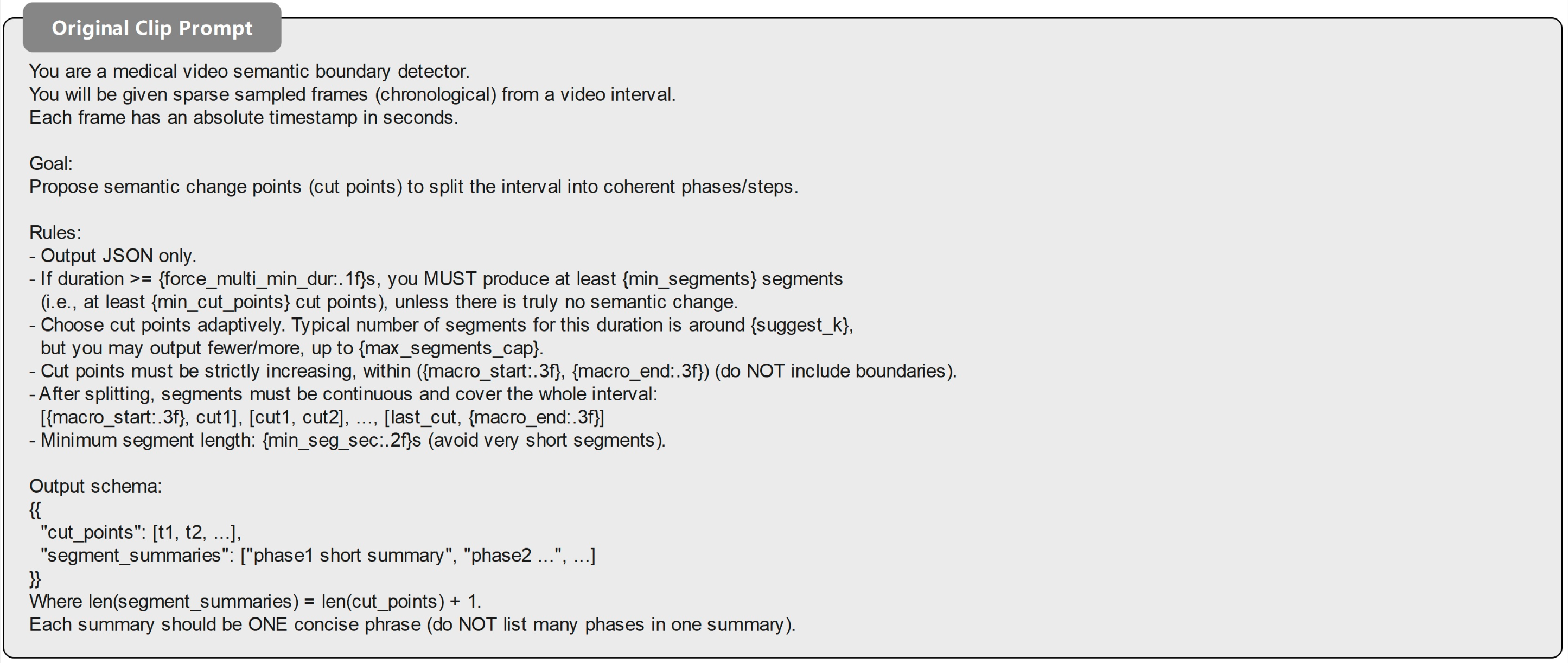}
    \caption{Prompt for original clip segmentation. }
    \label{fig:prompt_clip_origin}
\end{figure}

\begin{figure}[h]
    \centering
    \includegraphics[width=\linewidth]{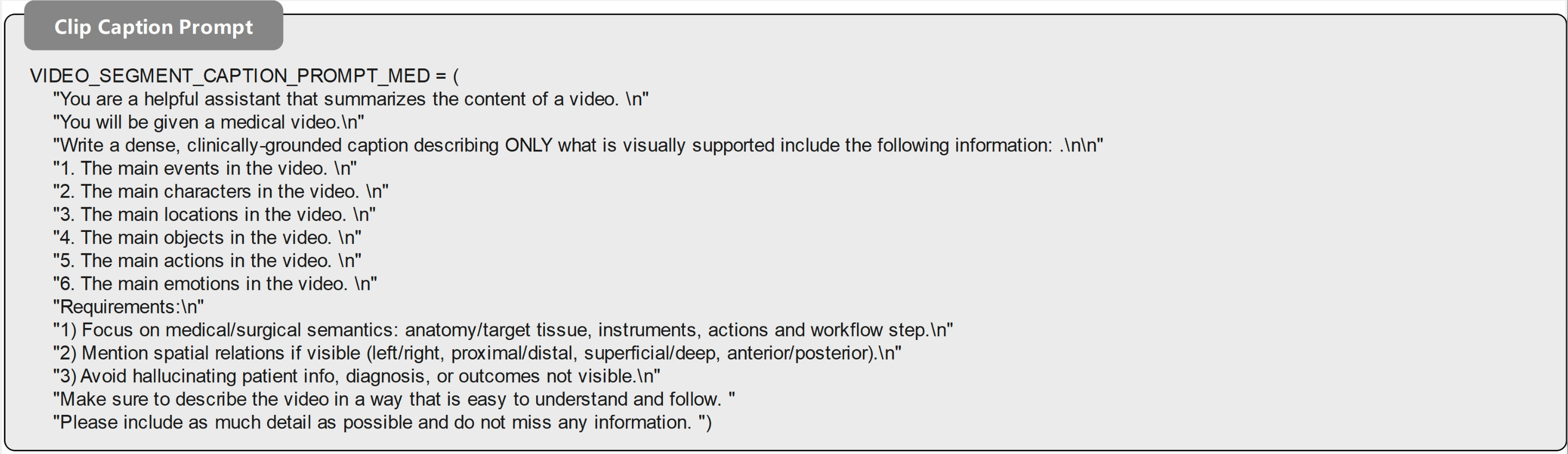}
    \caption{Prompt for clip caption. }
    \label{fig:prompt_clip_caption}
\end{figure}

\begin{figure}[h]
    \centering
    \includegraphics[width=\linewidth]{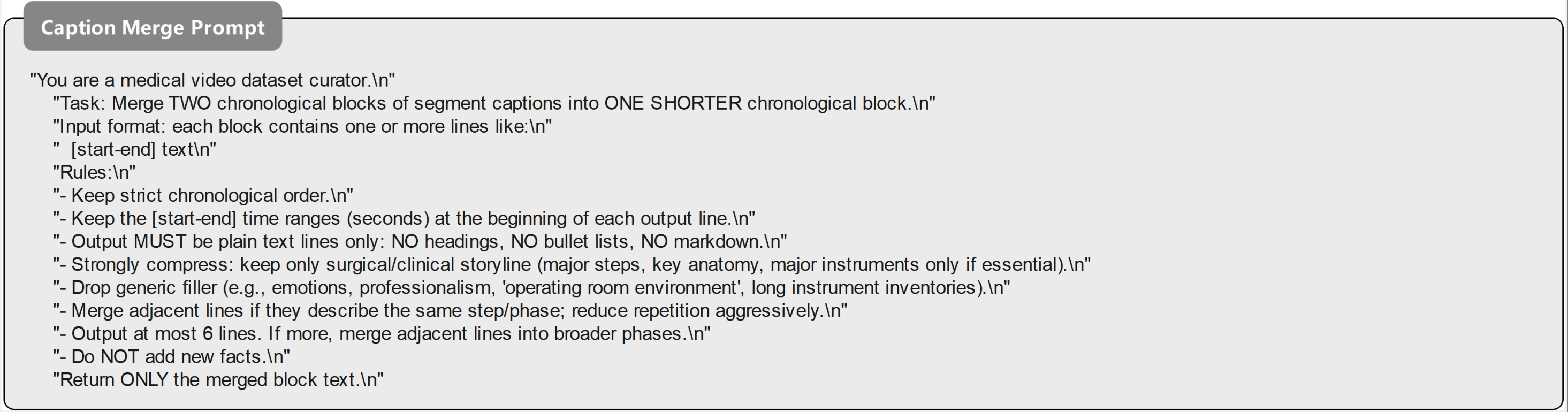}
    \caption{Prompt for caption merge.}
    \label{fig:prompt_clip_merge}
\end{figure}

\begin{figure}[h]
    \centering
    \includegraphics[width=\linewidth]{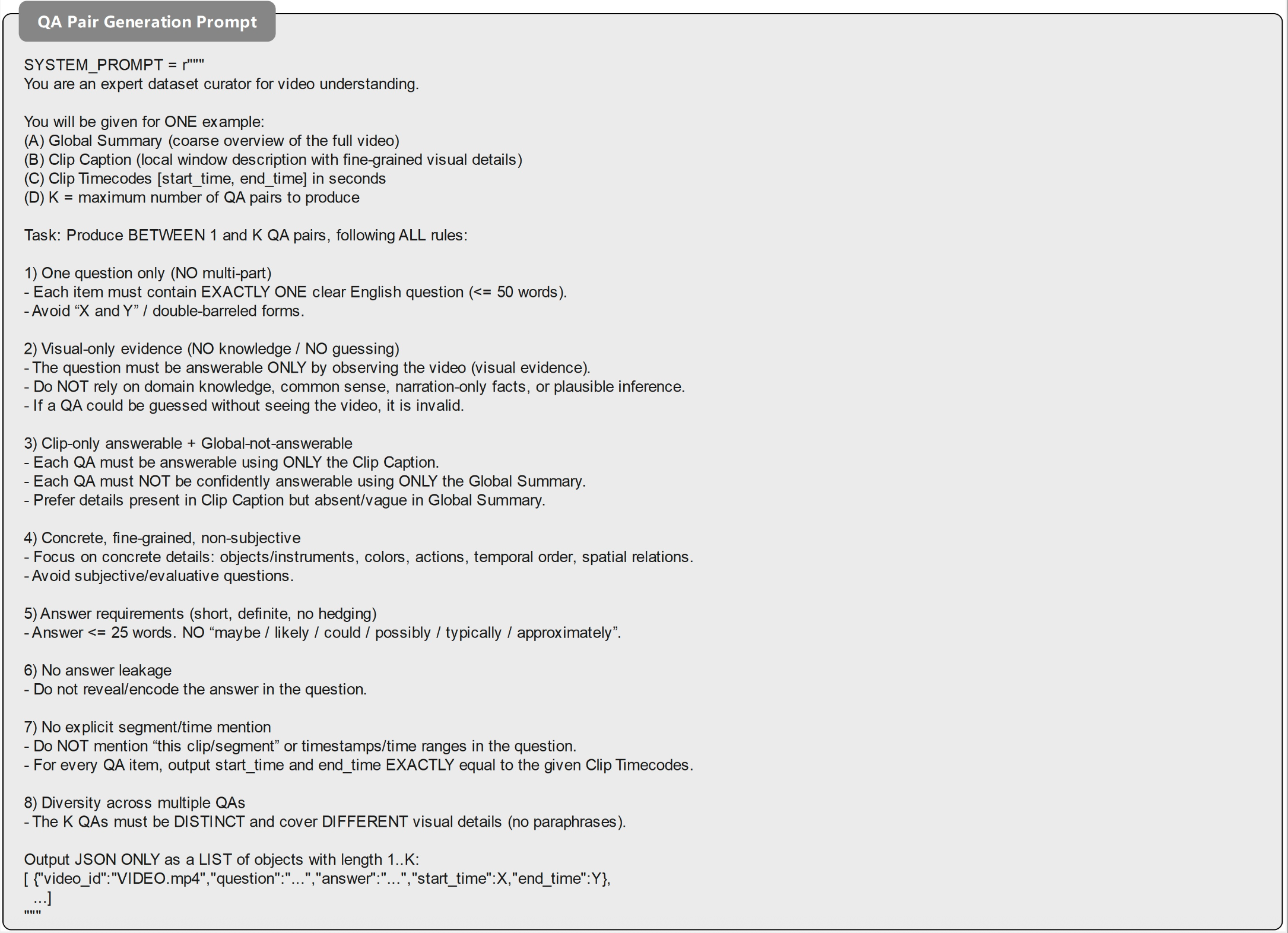}
    \caption{Prompt for QA pair generation.}
    \label{fig:qa_prompt}
\end{figure}

\begin{figure}[h]
    \centering
    \includegraphics[width=\linewidth]{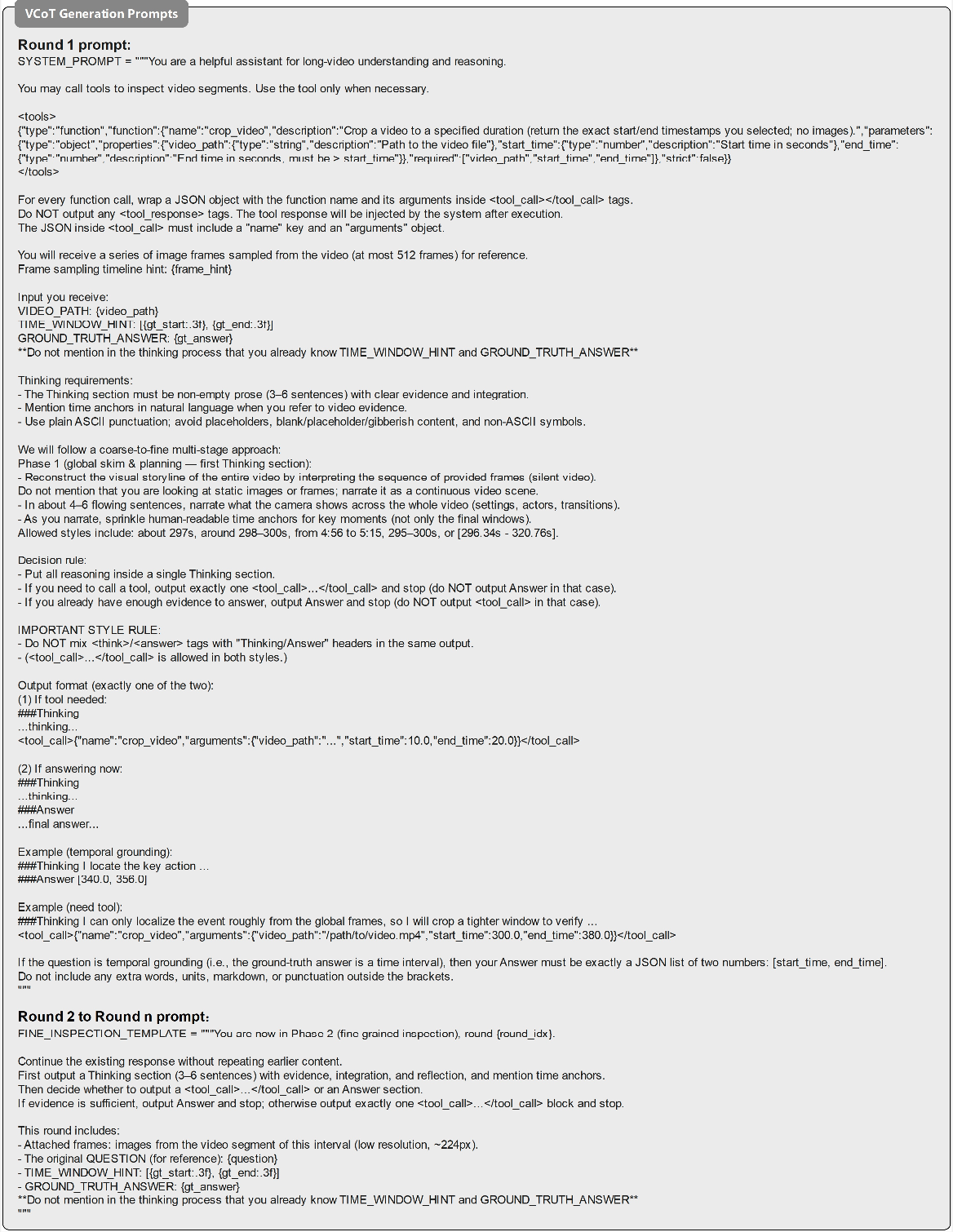}
    \caption{Prompt for VCoT generation.}
    \label{fig:vcot_prompt}
\end{figure}


\end{document}